%% file: main.tex
\PassOptionsToPackage{T2A,T1}{fontenc}
\documentclass{goodfire}
\usepackage{goodfire}
\usepackage{noto-sans}
\renewcommand{\sfdefault}{\cabinfamily}

\input{math_commands.tex}

\usepackage{hyperref}
\usepackage{cleveref}
\usepackage{url}
\usepackage{graphicx}
\usepackage{fontawesome5}
\usepackage{booktabs}
\usepackage{tabularx}
\usepackage{longtable}
\usepackage{wrapfig}
\usepackage{needspace}
\usepackage{placeins}
\usepackage{array}
\usepackage{CJKutf8}
\usepackage{fvextra}
\usepackage{amssymb}
\newcommand{\promptcyrillic}[1]{{\fontencoding{T2A}\fontfamily{cmtt}\selectfont#1}}
\DeclareUnicodeCharacter{0434}{\promptcyrillic{\cyrd}}
\DeclareUnicodeCharacter{0438}{\promptcyrillic{\cyri}}
\DeclareUnicodeCharacter{043A}{\promptcyrillic{\cyrk}}
\DeclareUnicodeCharacter{043E}{\promptcyrillic{\cyro}}
\DeclareUnicodeCharacter{043F}{\promptcyrillic{\cyrp}}
\DeclareUnicodeCharacter{0440}{\promptcyrillic{\cyrr}}
\newsavebox{\promptChineseTo}
\AtBeginDocument{\sbox{\promptChineseTo}{\scriptsize\begin{CJK}{UTF8}{gbsn}\CJKchar{"52}{"30}\end{CJK}}}
\DeclareUnicodeCharacter{5230}{\usebox{\promptChineseTo}}
\DeclareUnicodeCharacter{2192}{\ensuremath{\rightarrow}}
\DeclareUnicodeCharacter{23CE}{\ensuremath{\hookleftarrow}}
\DeclareUnicodeCharacter{25B8}{\ensuremath{\blacktriangleright}}
\usepackage{tikz}
\usetikzlibrary{arrows.meta,fit,positioning}

\usepackage[subdued,defaultmathsizes,italic,LGRgreek+]{mathastext}
\MTgreekfont{\notosansfamily}
\MTDeclareVersion{figurecaption}{T1}{\cabinfamily}{m}{n}
\DeclareCaptionFont{captionmath}{%
  \MTversion*{figurecaption}\let\bm\mathnormalbold
}
\AddToHook{env/tabular/begin}{\sffamily}
\AddToHook{env/tabular*/begin}{\sffamily}

\definecolor{baseprediction}{HTML}{1B6E3C}
\definecolor{ablatedprediction}{HTML}{A33A2B}
\newcolumntype{L}{>{\raggedright\arraybackslash}X}

\title{Large-scale Neuron Community Discovery Reveals Neuron-Sparse Behaviors in Language Models}

\authors{
  Igor Shilov\textsuperscript{1,2},
  Thomas Fel\textsuperscript{1},
  Tal Haklay\textsuperscript{1},
  Lucius Bushnaq\textsuperscript{1},
  Owen Lewis\textsuperscript{1}
}
\affiliations{
  \textsuperscript{1}Goodfire \quad
  \textsuperscript{2}Imperial College London
}

\abstract{
What reusable units of computation can we discover within a language model's neurons? Prior work has shown that individual forward passes can often be sparsified post-hoc to small groups of neurons, but it has been unclear whether these groups form coherent computational entities reused across inputs. We introduce \emph{neuron community discovery}, an unsupervised method for finding neuron groups that recur across prompts and are causally necessary and sufficient, within the layers they span, for common computations. The method combines gradient-based attribution, a graph connecting neurons that are repeatedly attributed together, and community detection. Applied to Llama-3.1-8B using ten million prompts, it discovers thousands of candidate communities. We evaluate them on held-out prompts through three causal tests: community necessity, sufficiency within a layer subset, and layer necessity. These tests identify groups whose removal disrupts predictions and whose retention preserves predictions despite ablating the remaining MLP neurons in the tested layers. Discovered communities support functions like grammatical agreement, token-probability calibration, fixed phrases, in-context copying, addition, and multiplication. With communities capped at 128 neurons across all layers, 10.3\% of held-out prompts have at least one layer's contribution explained on a neuron-sparse basis. Coverage remains 8.7\% when requiring at least four layers and 3.3\% when requiring at least twelve. These results show that small groups of existing neurons can provide reusable computational components across prompts, and that such groups can be discovered at scale.
\par\medskip
\noindent\href{https://github.com/goodfire-ai/neuron-communities}{\faGithub\enspace\textbf{github.com/goodfire-ai/neuron-communities}}\par
\noindent\href{https://goodfire.com/demos/neuron-communities}{\faGlobe\enspace\textbf{goodfire.com/demos/neuron-communities}}
}

\begin{document}

\maketitlebox

\section{Introduction}
\label{sec:introduction}
\input{sections/1_introduction}

\section{Related Work}
\label{sec:related-work}
\input{sections/2_related_work.tex}

\section{Method}
\label{sec:method}
\input{sections/4_method.tex}

\section{Examples}
\label{sec:examples}
\input{sections/5_examples.tex}

\section{Coverage}
\label{sec:coverage}
\input{sections/6_coverage.tex}

\section{Community deep dive}
\label{sec:deepdive}
\input{sections/7_deepdive.tex}

\section{Discussion and limitations}
\label{sec:discussion}
\input{sections/8_discussion.tex}

\section{Conclusion}
\label{sec:conclusion}

We introduced neuron community discovery, a method for finding small groups of neurons that support common computations without specifying the behaviors in advance. In Llama-3.1-8B, the method discovers thousands of communities, with groups of at most 128 neurons passing our necessity and sufficiency tests within at least one layer on 10.3\% of held-out prompts. Examples range from grammatical agreement and fixed phrases to copying and arithmetic. These results support small neuron groups as reusable units of computation in language models for at least some behaviors.

\clearpage
\bibliographystyle{plainnat}
\bibliography{bibliography}

\clearpage

\appendix
\crefalias{section}{appendix}
\crefname{appendix}{appendix}{appendices}
\Crefname{appendix}{Appendix}{Appendices}

\section{Recovering known arithmetic neurons}
\label{app:arithmetic-attribution}
\input{sections/appendix_arithmetic.tex}

\section{Attribution method ablation}
\label{app:attribution-ablation}
\input{sections/appendix_attribution_ablation.tex}

\section{Selecting resolution thresholds}
\label{app:gamma-sweep}
\input{sections/appendix_gamma_sweep.tex}

\section{Targeted data slices}
\label{app:targeted-data}
\input{sections/appendix_data.tex}

\FloatBarrier
\section{Behavioral taxonomy prompts}
\label{app:taxonomy-prompts}
\input{sections/appendix_taxonomy_prompt.tex}

\clearpage
\section{Additional community subset ablations}
\label{app:community-subsets}
\input{sections/appendix_community_subsets.tex}

\end{document}

%% file: math_commands.tex
\usepackage{amsmath,amsfonts,bm}

\def\eqref#1{equation~\ref{#1}}

\def\1{\bm{1}}

\DeclareMathAlphabet{\mathsfit}{\encodingdefault}{\sfdefault}{m}{sl}
\SetMathAlphabet{\mathsfit}{bold}{\encodingdefault}{\sfdefault}{bx}{n}



%% file: sections/1_introduction.tex
\begin{figure}[t]
    \centering
    \includegraphics[width=0.95\textwidth]{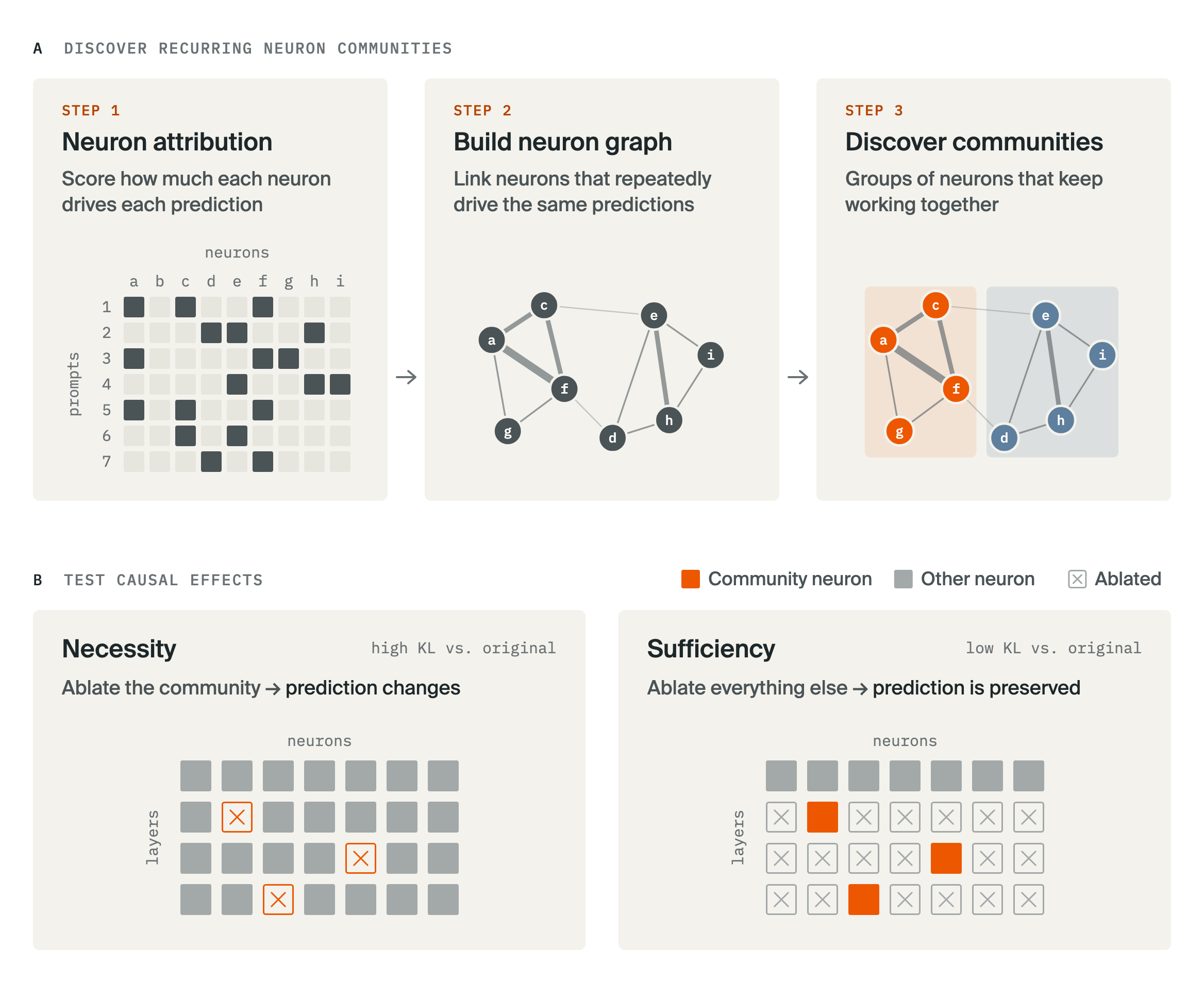}
    \caption{\textbf{Discovering and testing neuron communities.} \textbf{A:} We compute per-prompt neuron attributions, then construct a co-occurrence graph, and run the Leiden community detection algorithm. \textbf{B:} We test community necessity and sufficiency (within a subset of layers) for predictions on held-out prompts.}
    \label{fig:method-overview}
\end{figure}

What are the basic units of computation in a language model? We would like to identify components of the model that consistently implement a certain function across different prompts. Ideally, this decomposition would reflect the model's natural taxonomy of computations, and the building blocks would encode recurring patterns of model use. Existing techniques in interpretability make progress on parts of this goal. Sparse autoencoders (SAEs) learn a dictionary of reusable atoms directly from a heterogeneous corpus, but their training objective is to reconstruct a model's activations rather than to capture its computation. Per-layer transcoders approximate MLP computation within a single layer.

Attribution graphs built from cross-layer transcoder features trace computation across layers~\citep{ameisen2025circuit,lindsey2025biology}. Neuron-based circuits also show promise: \citet{arora2025neuron} show that neuron-based circuits can be as sparse and faithful as circuits built from sparse-autoencoder features on subject--verb agreement, and recover neuron groups involved in intermediate steps of multi-hop reasoning. \citet{feucht2026arithmetic} identify 28 neurons in layer 18 of Llama-3.1-8B that implement a base-10 addition mechanism reused across arithmetic tasks. Still, these analyses typically use small sets of human-selected prompts, implicitly assuming that the model's computations follow human task categories.

\looseness=-1
We reverse this approach: across millions of diverse prompts, we identify neuron groups repeatedly recruited together, then ask what computations they perform. This lets us discover recurring computational units without specifying their functions in advance.
We introduce \emph{neuron community discovery} to make this search scalable. We use gradient-based attribution to identify influential neurons in Llama-3.1-8B across ten million prompts from FineWeb-Edu, Wikipedia, and FineMath. We connect neurons repeatedly attributed together in a graph and use Leiden community detection at multiple resolution thresholds to find candidate \emph{communities} (\Cref{fig:method-overview}).
We evaluate these communities through causal interventions on held-out prompts. A community is \textbf{necessary} when ablating its neurons substantially changes the next-token distribution. It is \textbf{sufficient} when retaining its neurons, while ablating the other MLP neurons in the same set of layers, largely preserves the distribution.

As an example, one of the discovered communities is responsible for \emph{an} + vowel agreement: six neurons across six layers promote words starting with a vowel after the article \emph{an}. Removing these neurons substantially changes predictions on randomly sampled \emph{an}-article prompts, with a median KL divergence of 4.99. Retaining these six neurons and four ``always on'' neurons while removing the remaining MLP computation in those six layers largely preserves the prediction, with a median KL divergence of 0.053. This illustrates how a small neuron group distributed across several layers can be necessary and sufficient for a common linguistic behavior.

Our method discovers thousands of other communities with similar properties, supporting grammatical agreement, morphology, common expressions, formatting, in-context copying, and numerical operations. To measure how broadly these sparse explanations apply, we count unique held-out prompts for which at least one community passes causal tests. Pooling communities across six resolution thresholds covers 10.3\% of prompts within at least one layer. Coverage remains 8.7\% when we require at least four layers and 3.3\% when we require at least twelve, using communities of at most 128 neurons. The most common behaviors are grammatical and token-level, such as calibrating token probabilities or completing fixed phrases. Other communities support more complex behaviors, including in-context copying, addition, and multiplication.

These communities form a useful dictionary of reusable neuron groups, covering a substantial fraction of prompts. Many prompts remain unexplained, potentially reflecting computations that are not neuron-sparse as well as limitations of discovery.

%% file: sections/2_related_work.tex
\textbf{Circuits in language models.} Sparse autoencoders learn interpretable features from activations~\citep{bricken2023monosemanticity}, while transcoders approximate MLP computation with sparsely activating features~\citep{dunefsky2024transcoders}. Using native MLP neurons and relevance patching (RelP), \citet{arora2025neuron} recover subject--verb agreement circuits comparable in sparsity and faithfulness to circuits built from sparse-autoencoder features. They also identify neuron groups supporting intermediate computations in multi-hop reasoning, addition, and multilingual antonyms.

\textbf{Explaining individual neurons.} Prior work interprets MLP neurons as textual-pattern detectors that promote vocabulary items~\citep{geva2021memories}, and uses sparse probes to identify features represented by small neuron groups~\citep{gurnee2023haystack}. Other studies identify token and n-gram detectors~\citep{voita2024neurons}, neurons regulating prediction confidence and token-frequency preferences~\citep{stolfo2024confidence}, and neuron families shared across independently trained GPT-2 models~\citep{gurnee2024universal}. These findings relate to the token-level behaviors we discover.

\citet{cammarata2026clean} proposes that much apparent polysemanticity arises from recursive generalization across related patterns, providing the motivation for direct study of neurons and their circuits. \citet{tigges2026neurons} provide empirical evidence for sparse neuron computation in Qwen3.6-35B-A3B and use subcircuit analysis to characterize neurons' computational roles, with concise descriptions that predict and simulate strong activations.

\citet{nikankin2025arithmetic} identify sparse neuron sets implementing arithmetic heuristics, such as detecting operand ranges. \citet{feucht2026arithmetic} find 28 neurons in layer 18 of Llama-3.1-8B that perform base-10 addition in numerical tasks and questions about months, weekdays, and hours. These neurons group by Fourier features of different periods. Their reuse across tasks motivates our search for neuron groups supporting recurring behaviors.

\textbf{Automated circuit discovery.} Automated methods identify model components responsible for a behavior~\citep{conmy2023automated,kramar2024atp,marks2025sparse}. ACDC tests which connections can be removed without substantially changing a chosen behavior~\citep{conmy2023automated}. \citet{marks2025sparse} construct causal circuits from sparse-autoencoder features and introduce an unsupervised pipeline that discovers thousands of circuits and their associated behaviors, a close precedent for discovery at scale.

%% file: sections/4_method.tex
\paragraph{Overview.}

Given an input dataset, we first identify the top-$k$ most important neurons for each prompt using gradient-based attribution. We then discover communities of neurons that consistently co-occur in the top-$k$ sets across prompts. Finally, we test each neuron community causally on held-out data, asking whether the community is necessary and sufficient (within a certain subset of layers) for the prediction. \Cref{fig:method-overview} summarizes the pipeline.

\textbf{Step 1: Per-prompt neuron attribution.}
\label{sec:method:attribution}
For a prompt $x$, we compute neuron attributions only at the final token position of the prompt. Let $\hat y$ be the model's highest-probability next token and let $h_{\ell j}(x)$ denote the post-nonlinearity activation of MLP neuron $j$ in layer $\ell$ at this position.
We score the neuron's first-order influence on the prediction as
\begin{equation}
    a_{\ell j}(x)
    = \left|h_{\ell j}(x)\frac{\partial \log p(\hat y\mid x)}{\partial h_{\ell j}(x)}\right|.
    \label{eq:activation-gradient-attribution}
\end{equation}

We compute $a_{\ell j}$ for all neurons across all layers. For each layer, we retain the top-$k$ highest-scoring neurons and discard the rest, yielding a sparse attributed set $A(x)$ for every prompt.

We validate this attribution method by recovering known arithmetic neurons: the top 40 neurons ranked by mean attribution for arithmetic prompts recover 60.7--93.3\% of the neurons found by \citet{feucht2026arithmetic} (\Cref{app:arithmetic-attribution})  by learning causal subspaces on predefined arithmetic tasks and selecting neurons whose output weights align with those subspaces. We quantify attribution quality and compare with alternative neuron attribution scores and random baselines in \Cref{app:attribution-ablation}.

\textbf{Step 2: Co-occurrence graph.}
\label{sec:method:communities}
We build an undirected graph whose nodes are MLP neurons. For each pair of nodes, we count how many prompts attribute both neurons. Raw co-occurrence is confounded by neuron popularity, since frequently attributed neurons will co-occur often even without a specific relationship. We therefore compute a z-score measuring how unusually high this count is given the nodes' individual frequencies. We use this z-score as the edge weight and retain pairs that meet minimum count and z-score thresholds (see ``Experimental Setup'' below).

\textbf{Step 3: Community discovery.}
We use Leiden community detection~\citep{traag2019louvain} with the Constant Potts Model objective, which allows us to search for small neuron communities at multiple resolution thresholds~\citep{traag2011narrow}.
The resolution parameter $\gamma$ controls granularity: smaller values produce larger communities, while larger values split the graph into finer groups.
Because no single resolution is appropriate for every mechanism, we run Leiden over a grid of $\gamma$ values and pool the resulting communities together.
Additionally, we identify a set of ``always on'' neurons: those present in the top-$k$ most important neurons for more than 20\% of prompts.

\textbf{Step 4: Testing necessity and sufficiency.}
\label{sec:method:certification}

We are looking for prompts for which a community is necessary and sufficient to preserve the model's prediction. For each candidate $(x,C)$, we test community necessity, sufficiency, and layer necessity through interventions at the prediction position, measuring KL divergence from the clean next-token distribution $p_x$ to the intervened distribution. We test every subspan of the community's ordered layer list, allowing communities to explain predictions within all or only some of the layers they span. $S$ denotes a subset of the layers containing community neurons (not necessarily contiguous).

To test \emph{community necessity}, we zero-ablate all community neurons within the subspan ($C_S = C \cap S$), producing the distribution $p_x^{-C_S}$. We measure:
\begin{equation}
    \operatorname{Nec}(x,C,S)=D_{\mathrm{KL}}\!\left(p_x\,\|\,p_x^{-C_S}\right).
    \label{eq:community-necessity}
\end{equation}

To test \emph{sufficiency}, we zero-ablate all MLP neurons in $S$ except the community and ``always on'' neurons, producing the distribution $p_x^{C\mid S}$. We measure:
\begin{equation}
    \operatorname{Suff}(x,C,S)=D_{\mathrm{KL}}\!\left(p_x\,\|\,p_x^{C\mid S}\right).
    \label{eq:community-sufficiency}
\end{equation}

To make sure that the sufficiency measurement is meaningful, we test \emph{layer necessity}, i.e. whether removing a full layer set $S$ including community $C$ distorts the prediction. To test layer necessity, we zero-ablate all MLP neurons in $S$ except the ``always on'' neurons, producing the distribution $p_x^{-S}$. We measure:
\begin{equation}
    \operatorname{LayerNec}(x,S)=D_{\mathrm{KL}}\!\left(p_x\,\|\,p_x^{-S}\right).
    \label{eq:layer-necessity}
\end{equation}

A community $C$ is then considered \emph{necessary and sufficient} on a given prompt $x$ and within a layer set $S$ when $\operatorname{Nec}(x,C,S)\geq\tau_{\mathrm{N}}$, $\operatorname{LayerNec}(x,S)\geq\tau_{\mathrm{N}}$, and $\operatorname{Suff}(x,C,S)\leq\tau_{\mathrm{S}}$.

We test communities only on prompts where at least one member is in the attributed top-$k$.

\textbf{Experimental Setup.}
\label{sec:method:setup}
We run the discovery algorithm on Llama-3.1-8B~\citep{grattafiori2024llama3} on 10 million prompts from the data mix: 50\% FineWeb-Edu~\citep{penedo2024fineweb}, 25\% English Wikipedia~\citep{wikidump}, 25\% FineMath~\citep{benallal2025smollm2}. We additionally include a curated list of smaller sources (500k prompts total), targeting behaviors known to have neuron-sparse representations (see~\Cref{app:targeted-data}). For necessity and sufficiency evaluations, we use a further 10 million prompts from the same data mix.

We retain the top-$k$ neurons per layer with $k=5$ and construct the discovery graph over layers 0--29. We require a co-occurrence count of at least 5 and a z-score of at least 8. We exclude layers 30--31 because their attributions appear less neuron-sparse even for individual prompts, making community discovery noisier. We run Leiden community discovery over the six resolutions $\gamma\in\{32,64,128,256,512,1024\}$ (see \Cref{app:gamma-sweep} for the resolution threshold sweep motivating this choice). We remove 17 ``always on'' neurons after partitioning and retain communities of up to 128 neurons. We use $\tau_{\mathrm{N}}=0.7$ and $\tau_{\mathrm{S}}=0.2$.

%% file: sections/5_examples.tex
\begin{figure}[t]
    \centering
    \subfloat[]{\label{fig:an1}\includegraphics[width=0.55\textwidth]{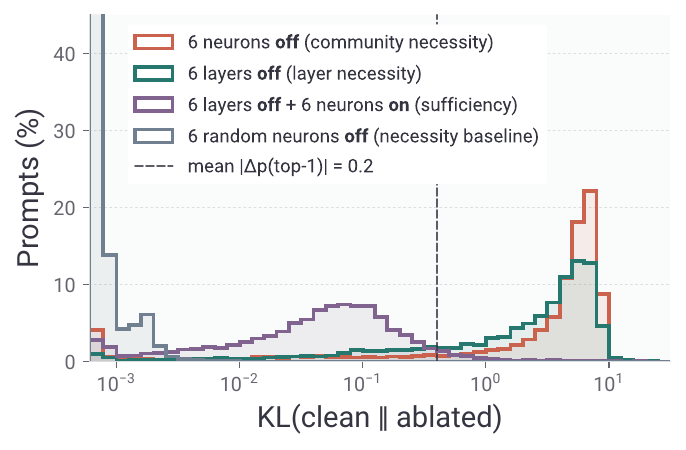}}
    \quad
    \subfloat[]{\label{fig:an2}\includegraphics[width=0.4\textwidth]{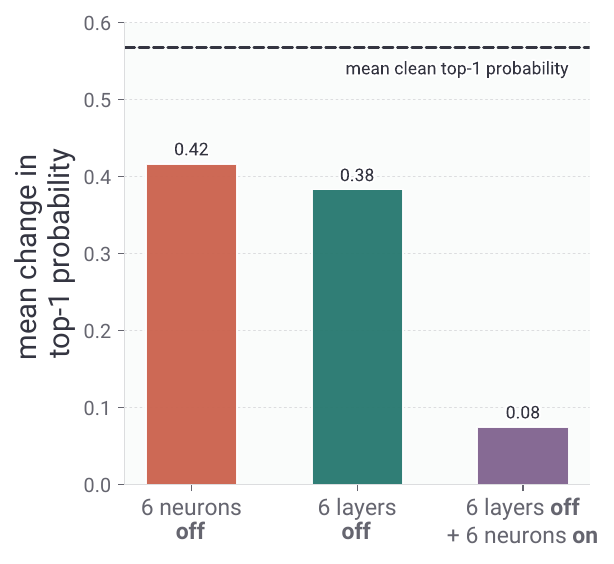}}
    \caption{\textbf{Six neurons across six layers implement \emph{an} + vowel agreement.} We show the community of six neurons to be necessary and sufficient to implement \emph{an} + vowel agreement. On a random set of prompts ending in the article \emph{an}, ablating the community distorts the prediction for most prompts (orange), while keeping only community and ``always on'' neurons reconstructs the original prediction well (purple). Ablating the six involved layers is shown in teal. Panel (a) also includes a necessity baseline obtained by ablating six randomly selected neurons (gray). We quantify the effects using: \textbf{(a)} KL divergence from the clean prediction. \textbf{(b)} Mean absolute change in the probability assigned to the clean top-1 token. The dashed line on the left represents the KL divergence threshold corresponding to a mean absolute change of 0.2 in the clean top-1 token's probability.}
    \label{fig:an}
\end{figure}

\subsection{Case study: an + vowel agreement}
\label{sec:an-case-study}

We first examine one of the discovered communities responsible for \emph{an} + vowel agreement. Of the 20,339 prompts on which the full discovered community passes the necessity and sufficiency tests, 99.7\% end at an \emph{an} token, always used as an article, rather than as a fragment within a word. Below we use ``community \emph{active on} a prompt'' as shorthand to mean that a community passes sufficiency and necessity tests on that prompt.

To assess the community's recall for this behavior, we sample 8,000 prompts ending in the article token \texttt{\textvisiblespace an} independently of whether the community is active. We focus on the six community neurons in the L2--L14 subspan: \texttt{\textbf{L2}:5664}, \texttt{\textbf{L5}:3846}, \texttt{\textbf{L8}:13390}, \texttt{\textbf{L12}:14182}, \texttt{\textbf{L13}:3604}, and \texttt{\textbf{L14}:11888}. We test their necessity, sufficiency, and layer necessity, intervening only in these six layers and retaining the four ``always on'' neurons (\texttt{\textbf{L2}:4786}, \texttt{\textbf{L5}:7012}, \texttt{\textbf{L12}:13860}, \texttt{\textbf{L13}:4208}).

Removing the six community neurons substantially changes the model's predictions (orange in \Cref{fig:an}). The median KL divergence is 4.99. The mean absolute change in the probability of the clean top-1 token is 0.42, against a mean clean probability of 0.57. Removing all MLP neurons in the same six layers except the four ``always on'' neurons has a comparable effect (teal in \Cref{fig:an}): the median KL divergence is 3.15, and the mean absolute probability change is 0.38. By comparison, ablating six randomly selected neurons leaves most predictions nearly unchanged, with KL divergence concentrated near zero (gray in \Cref{fig:an1}).

In contrast, keeping only the six community neurons, out of 86,016 total in those layers, along with the four ``always on'' neurons, largely preserves the prediction (purple in \Cref{fig:an}). The KL distribution peaks below 0.1, with a median of 0.053. The mean absolute change in the clean top-1 token's probability falls to 0.08. Thus, on most an-article prompts, these six neurons preserve the prediction despite removal of the remaining MLP computation in the corresponding six layers.

The small peak near zero KL under community ablation identifies prompts that are largely unaffected by the ablation, e.g. common expressions such as ``to give you an [example]''.

We examine how predictions change as increasing subsets of community neurons are removed in \Cref{sec:deepdive}. The community seem to operate mostly as a \emph{redundant voting} of individual neurons: removing any one neuron does almost no damage, but removing several at once can starts to substantially disrupt the prediction.

\input{sections/table_community_examples.tex}

\subsection{Representative behaviors}
\label{sec:representative-behaviours}

The \emph{an} + vowel community is one of thousands discovered by our method. \Cref{tab:community-examples} illustrates their range of behaviors, spanning grammar, morphology, formatting, in-context lookup, numerical operations, and token-level operations. For each example, we report neuron and layer counts within the selected subspan, a representative prompt, and predictions before and after community ablation. An interactive explorer of the full collection is available at \url{https://goodfire.com/demos/neuron-communities}.

\looseness=-1
Many discovered communities perform grammatical and morphological roles. A community of nine neurons across nine layers selects \emph{is} or \emph{are} after \emph{there}: ablating it changes the prediction after ``Ten cards are put into five envelopes so that there'' from \emph{are} to \emph{is}. Another community detects the plural suffix \emph{-s}; removing it makes the model behave as though the preceding word were singular. Formatting communities support operations such as completing a LaTeX command or closing a parenthesis.

Other communities support copying information from the context. \Cref{tab:community-examples} shows how ablating seven neurons prevents the model from retrieving the filename (\emph{Delta.exe}) mentioned in the context and leads it to produce a generic continuation instead. Induction heads provide a known mechanism for copying continuations from context~\citep{olsson2022induction}. The community may work alongside such heads, although we have not tested that interaction.

Numerical communities support more complex computations. We recover an eight-neuron addition community that is necessary and sufficient on a range of addition prompts. A multiplication community supports multiplication and division across several settings: explicit arithmetic such as ``2*8='', the prime-factorization example in \Cref{tab:community-examples}, and unit conversion, e.g. a model predicting ``68mi'' following ``112km (''.

Many behaviors are token-level operations, including suppressing particular tokens and promoting common expressions such as \emph{each other} or \emph{made up}. These expression communities are sensitive to the intended continuation: the \emph{each other} community promotes \emph{each} only when it is used as a part of the phrase, but not in other contexts. Another community consistently reduces the probability of the token \emph{fraction}.

\begin{figure}[t]
    \centering
    \includegraphics[width=0.95\textwidth]{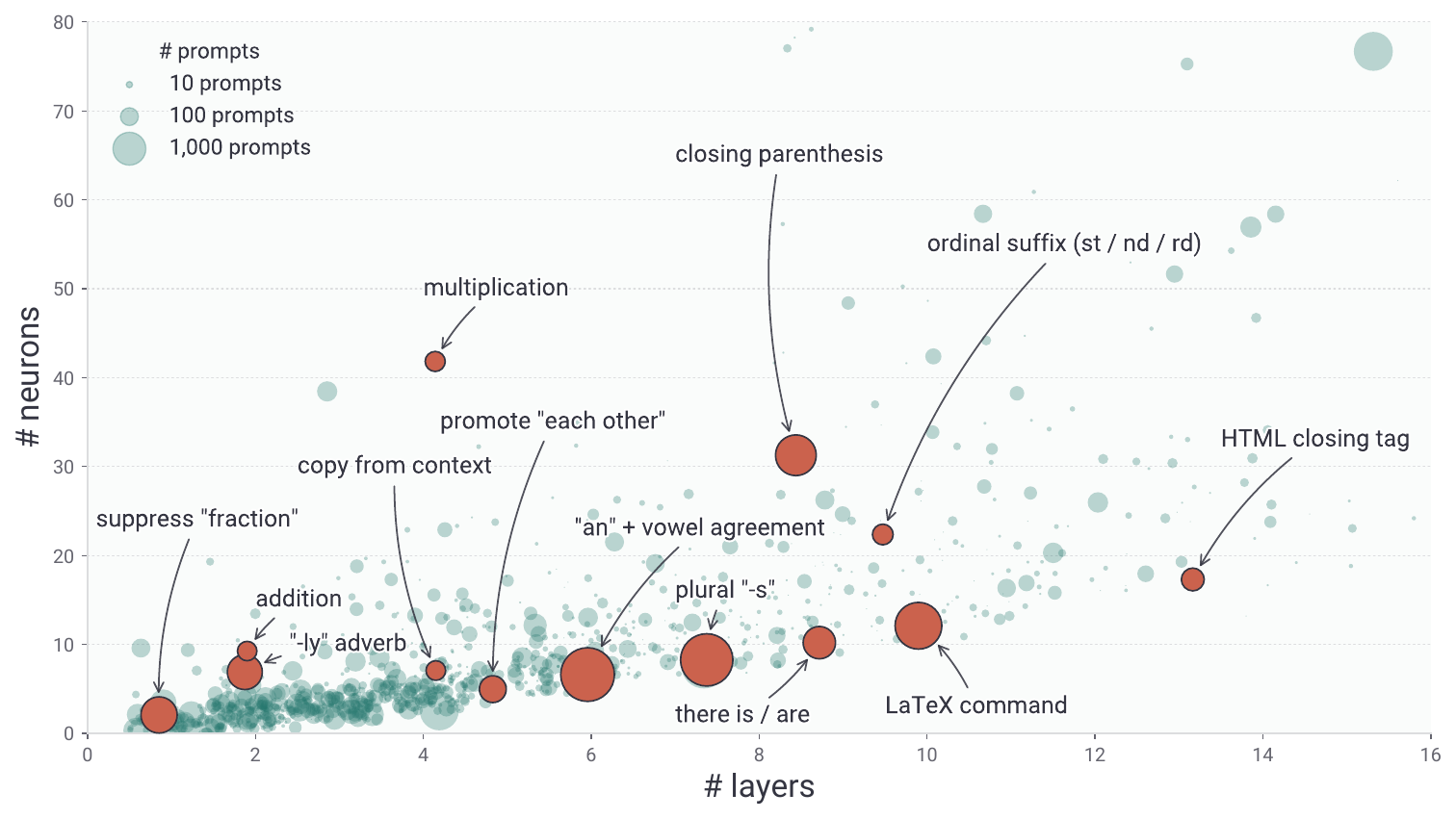}
    \caption{\textbf{Neuron and layer counts for discovered communities.} Teal points show a random sample of 1,000 communities that explain at least ten prompts. Orange points highlight representative behaviors from~\Cref{tab:community-examples}. Marker size indicates the number of prompts explained.}
    \label{fig:fig1}
\end{figure}

Most communities we inspect have similarly consistent effects across their associated prompts.
\Cref{fig:fig1} shows a broader view of community size and layer distribution. Each point represents a community, with the y-axis representing the number of neurons in that community and the x-axis representing the number of layers these neurons come from. A larger circle indicates a community with more prompts for which it passes the causal tests. The figure shows a subsample of 1,000 communities, randomly sampled at the prompt level. The densest region contains communities with fewer than ten neurons across two or three layers, often supporting token-level operations and common expressions. Other displayed communities extend across roughly 15 layers, showing that a small neuron set can participate in computations distributed across much of the model.

%% file: sections/table_community_examples.tex
\begin{table}[t]
    \newcommand{\nexttokenmarker}{\textcolor{gray}{\textvisiblespace}}
    \caption{\textbf{Example neuron communities and their causal effects on model predictions}.}
    \label{tab:community-examples}
    \centering
    \small
    \setlength{\tabcolsep}{4pt}
    \renewcommand{\arraystretch}{1.15}
    \newcolumntype{E}{
        >{\fontsize{8}{9.5}\selectfont\raggedright\arraybackslash}p{0.16\textwidth}
        >{\fontsize{8}{9.5}\selectfont\centering\arraybackslash}p{0.09\textwidth}
        >{\fontsize{8}{9.5}\selectfont\raggedright\arraybackslash}p{0.15\textwidth}
        >{\fontsize{8}{9.5}\selectfont}L
        >{\fontsize{8}{9.5}\selectfont\raggedright\arraybackslash}p{0.20\textwidth}}
    \begin{tabular}{@{}>{\centering\arraybackslash}m{12pt}@{\hspace{6pt}}l@{}}
        \toprule
        &
        \begin{tabularx}{\dimexpr\textwidth-18pt\relax}{@{}E@{}}
        \small\textbf{Function}
            & \small\textbf{Neurons}
            & \small\textbf{Layers}
            & \small\textbf{Prompt}
            & \small\textbf{Predictions}\newline
              (\textcolor{baseprediction}{\textbf{base}}, \textcolor{ablatedprediction}{\textbf{ablated}}) \\
        \end{tabularx} \\
        \midrule
        \rotatebox{90}{\scriptsize\textbf{Grammar}} &
        \begin{tabularx}{\dimexpr\textwidth-18pt\relax}{@{}E@{}}
        Promote vowel-initial words after \emph{an}
            & 6
            & 6 $\subseteq$ L2--L14
            & Diagnostic services play an~\nexttokenmarker
            & \textcolor{baseprediction}{essential (0.75)}\newline
              \textcolor{ablatedprediction}{crucial (0.33)} \\
        \midrule
        Select \emph{is} or \emph{are} after \emph{there}
            & 9
            & 9 $\subseteq$ L0--L29
            & Ten cards are put into five envelopes so that there~\nexttokenmarker
            & \textcolor{baseprediction}{are (1.00)} \newline
              \textcolor{ablatedprediction}{is (0.65)} \\
        \end{tabularx} \\
        \midrule
        \rotatebox{90}{\scriptsize\textbf{Morphology}} &
        \begin{tabularx}{\dimexpr\textwidth-18pt\relax}{@{}E@{}}
        Detect plural \emph{-s}
        & 9
        & 7 $\subseteq$ L0--L6
        & Finally, some NGOs~\nexttokenmarker
        & \textcolor{baseprediction}{are (0.13)}\newline
        \textcolor{ablatedprediction}{leaders (0.53)} \\
        \midrule
        Detect adverbial \emph{-ly}
        & 8
        & 2 $\subseteq$ L0--L1
        & two independent normally~\nexttokenmarker
        & \textcolor{baseprediction}{distributed (1.00)}\newline
        \textcolor{ablatedprediction}{variables (0.99)} \\
        \end{tabularx} \\
        \midrule
        \rotatebox{90}{\scriptsize\textbf{Formatting}} &
        \begin{tabularx}{\dimexpr\textwidth-18pt\relax}{@{}E@{}}
        Complete LaTeX commands
        & 11
        & 10 $\subseteq$ L0--L25
        & sampling distribution of \texttt{\$\textbackslash bar\{y\}\_1 - \textbackslash}~\nexttokenmarker
        & \textcolor{baseprediction}{bar (1.00)} \newline
        \textcolor{ablatedprediction}{\texttt{\textbackslash} (0.94)} \\
        \midrule
        Close parentheses
        & 32
        & 8 $\subseteq$ L0--L8
        & \texttt{class Solution(object}~\nexttokenmarker
        & \textcolor{baseprediction}{\texttt{):} (0.96)} \newline
        \textcolor{ablatedprediction}{\texttt{:} (0.92)} \\
        \end{tabularx} \\
        \midrule
        \rotatebox{90}{\scriptsize\textbf{In-context}} &
        \begin{tabularx}{\dimexpr\textwidth-18pt\relax}{@{}E@{}}
        Copy an entity from context
        & 7
        & 4 $\subseteq$ L23--L27
        & malware camouflages as Delta.exe ... should check the~\nexttokenmarker
        & \textcolor{baseprediction}{Delta (1.00)}\newline
        \textcolor{ablatedprediction}{location (0.85)} \\
        \midrule
        Complete HTML closing tags
        & 17
        & 13 $\subseteq$ L0--L28
        & \texttt{<div>JavaScript is disabled on your browser.</}~\nexttokenmarker
        & \textcolor{baseprediction}{\texttt{div} (1.00)}\newline
        \textcolor{ablatedprediction}{\texttt{</} (1.00)} \\
        \end{tabularx} \\
        \midrule
        \rotatebox{90}{\scriptsize\textbf{Numerical operations}} &
        \begin{tabularx}{\dimexpr\textwidth-18pt\relax}{@{}E@{}}
        Multiplication
        & 42
        & 4 $\subseteq$ L18--L21
        & Prime factorization: 799 = 17 x~\nexttokenmarker
        & \textcolor{baseprediction}{47 (1.00)}\newline
        \textcolor{ablatedprediction}{43 (0.51)} \\
        \midrule
        Addition
        & 8
        & 2 $\subseteq$ L18--L19
        & are 43 active cases (40 at the border and~\nexttokenmarker
        & \textcolor{baseprediction}{three (0.88)}\newline
        \textcolor{ablatedprediction}{four (0.65)} \\
        \midrule
        Select ordinal suffix (\emph{st/nd/rd})
        & 21
        & 9 $\subseteq$ L1--L27
        & Last updated date: 22~\nexttokenmarker
        & \textcolor{baseprediction}{nd (1.00)}\newline
        \textcolor{ablatedprediction}{th (0.99)} \\
        \end{tabularx} \\
        \midrule
        \rotatebox{90}{\scriptsize\textbf{Token-level}} &
        \begin{tabularx}{\dimexpr\textwidth-18pt\relax}{@{}E@{}}
        Suppress \emph{fraction}
        & 2
        & 1 $\subseteq$ L29--L29
        & we could conceivably give~\nexttokenmarker
        & \textcolor{baseprediction}{fractions (0.21)}\newline
        \textcolor{ablatedprediction}{fractions (0.74)} \\
        \midrule
        Promote \emph{each other}
        & 6
        & 5 $\subseteq$ L12--L29
        & farmers often come together to help~\nexttokenmarker
        & \textcolor{baseprediction}{each (0.81)}\newline
        \textcolor{ablatedprediction}{their (0.67)} \\
        \end{tabularx} \\
        \bottomrule
    \end{tabular}
\end{table}

%% file: sections/6_coverage.tex
\subsection{Prompt coverage}
\label{sec:prompt-coverage}

We now quantify how often the discovered communities explain the model's predictions. We run graph partitioning at six resolution thresholds, $\gamma\in\{32,64,128,256,512,1024\}$, each producing its own set of communities. These sets can overlap in both their neurons and the behaviors they support, but searching multiple thresholds recovers communities that are missed at any single threshold.

\begin{figure}[t]
    \centering
    \subfloat[]{\label{fig:cov1}\includegraphics[width=0.3\textwidth]{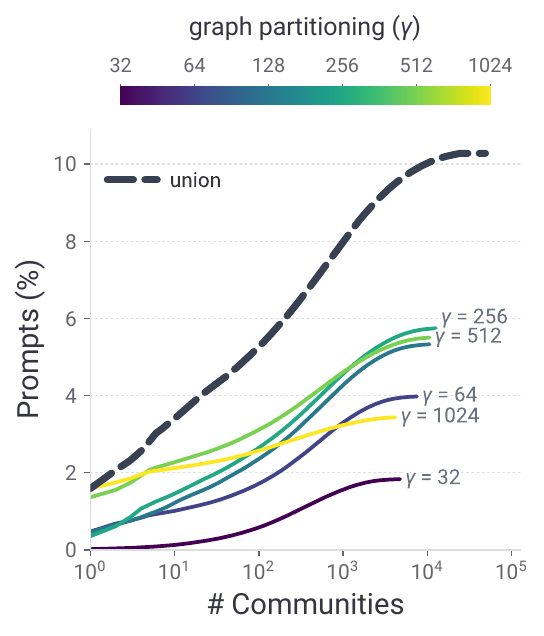}}
    \quad
    \subfloat[]{\label{fig:cov2}\includegraphics[width=0.3\textwidth]{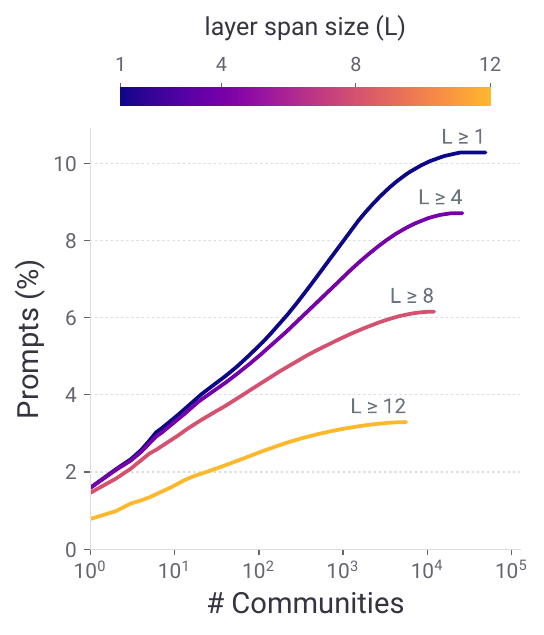}}
    \quad
    \subfloat[]{\label{fig:cov3}\includegraphics[width=0.3\textwidth]{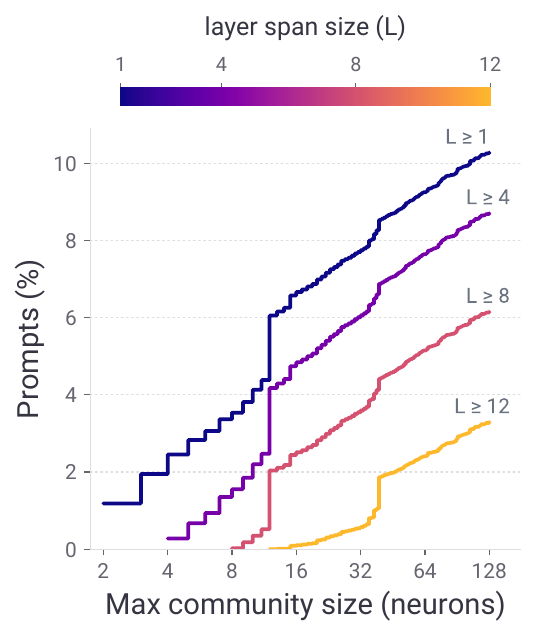}}
    \caption{\textbf{Neuron-sparse communities explain 10.3\% of held-out prompts.} The y-axis shows the proportion of prompts explained by sparse neuron communities.
    \textbf{(a)} Communities discovered at a given resolution threshold $\gamma$ explain the behavior of at least one layer for up to 5.7\% of prompts. Coverage rises to 10.3\% when we take the union of communities discovered across the six thresholds.
    \textbf{(b)} For each prompt, we require a single community to explain at least $L$ layers. Coverage is 10.3\% for $L=1$, 8.7\% for $L=4$, and 3.3\% for $L=12$.
    \textbf{(c)} Varying the per-community sparsity definition, we show that even with at most 16 neurons per community, 6.7\% of prompts have at least one layer explained, and 2.5\% of prompts have at least 8 layers explained.
    }
    \label{fig:cov}
\end{figure}

Under the least restrictive definition, a prompt is \textit{explained} by a neuron-sparse community if at least one layer's contribution can be explained by community neurons, with all three causal tests passing: necessity, sufficiency, and layer necessity. Here, ``explained'' is shorthand for a partial explanation of the prediction, where a community explains the contribution of certain layers to the final model prediction. \Cref{fig:cov1} shows the proportion of unique prompts covered for each resolution threshold and for their union. Communities are added greedily, choosing the community that explains the most previously uncovered prompts at each step. At $\gamma=256$, only ten communities cover approximately 1.5\% of prompts, including behaviors such as the \emph{an} + vowel agreement examined in \Cref{sec:an-case-study}. Pooling communities across all six thresholds increases coverage to 10.3\%: for roughly one in ten prompts, we find a neuron-sparse community that is necessary and sufficient within at least one layer -- the random baseline only achieves ~3\% in the same setup ((\Cref{app:attribution-ablation})).

\Cref{fig:cov2} shows the coverage for explanations involving several layers. A prompt counts at threshold $L$ if a single community found at any $\gamma$ passes all three causal tests within the same set of at least $L$ distinct layers (which need not be contiguous). Coverage remains high at 8.7\% for $L\geq4$ and 3.3\% for $L\geq12$. Thus, individual sparse communities can explain contributions from several layers simultaneously.

In the experiments so far, we have set a sparsity threshold of 128 neurons across all layers. \Cref{fig:cov3} varies this arbitrary threshold, while retaining the same causal criteria. Communities with at most 16 neurons already explain at least one layer on approximately 6.7\% of prompts, showing that substantial coverage persists under a much tighter neuron budget.

\subsection{Behavioral taxonomy}
\label{sec:coverage-taxonomy}

\begin{wraptable}{r}{0.5\textwidth}
    \centering
    \small
    \setlength{\tabcolsep}{3pt}
    \renewcommand{\arraystretch}{1.15}
    \vspace{-.9cm}
    \caption{Frequency of neuron-sparse behaviors and associated communities, as classified by an auto-interp pipeline with \texttt{gpt-5.6-terra}.}
    \begin{tabularx}{\linewidth}{@{}Lrr@{}}
        \toprule
        \textbf{Class} & \shortstack[r]{\textbf{Communities}} & \shortstack[r]{\textbf{Prompts}} \\
        \midrule
        Common phrases & 42.3\% & 31.3\% \\
        Token suppression & 19.5\% & 19.6\% \\
        Formatting & 9.3\% & 19.4\% \\
        Grammar & 2.9\% & 15.3\% \\
        Sub-word completion & 8.0\% & 6.0\% \\
        Topic & 9.0\% & 3.0\% \\
        Retrieval and computation & 3.7\% & 2.9\% \\
        Other & 5.3\% & 2.5\% \\
        \bottomrule
    \end{tabularx}
    \label{tab:coverage-taxonomy}
\end{wraptable}

To characterize the behaviors represented by the discovered communities, we use an LLM to describe and classify communities found at $\gamma=256$ that pass the causal tests on at least 10 held-out prompts. For each community, we provide \texttt{gpt-5.6-terra} with up to 100 prompts where the community is necessary and sufficient, together with the model's predictions before and after community ablation, and the ground truth. The LLM assigns one primary label from a manually created taxonomy of eight broad classes in \Cref{tab:coverage-taxonomy}. \Cref{app:taxonomy-prompts} provides the classifier prompt, taxonomy, and an example input.

Lexical behaviors, such as completing common phrases like \emph{make up}, form the largest class, accounting for 42.3\% of communities and covering 31.3\% of prompts explained by at least one classified community. Per-token probability calibration, such as suppressing a fixed word, e.g. \emph{fraction}, accounts for another 19.5\% of communities. Community counts and prompt coverage differ substantially across classes: grammar and agreement, such as \emph{an} + vowel agreement or singular/plural agreement, comprise only 2.9\% of communities but cover 15.3\% of these covered prompts. Topic-related communities, such as those promoting vocabulary about football or abstract algebra, comprise 9.0\% of communities and cover 3.0\% of prompts. These results distinguish frequently discovered behaviors from those that apply broadly across prompts.

\begin{figure}[t]
    \centering
    \includegraphics[width=0.95\textwidth]{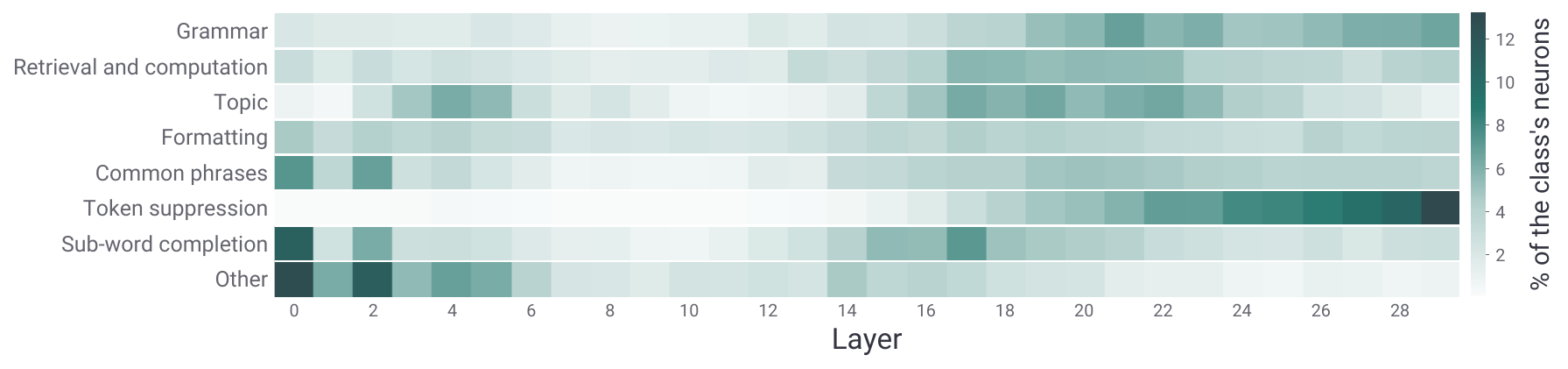}
    \caption{\textbf{Different behaviors rely on neurons at different depths.} Each row shows the distribution across layers of neurons in communities assigned to a given behavioral class.}
    \label{fig:taxonomy_heatmap}
\end{figure}

\Cref{fig:taxonomy_heatmap} shows how the neurons in each behavior are distributed across layers. Prior work also finds differences with depth in the textual patterns detected by MLP neurons~\citep{geva2021memories}.
Per-token suppression/promotion, for instance, is concentrated in the latest layers. Sub-word completion peaks at L0, with a secondary peak around L15--L17. Grammar is more concentrated from L18 onward, while retrieval and computation, including numerical operations, peak around L17--L22.

%% file: sections/7_deepdive.tex
To examine how neurons within a discovered community work together, we return to the six-neuron \emph{an} + vowel community from \Cref{sec:an-case-study}, using 100 randomly sampled prompts on which it passes the necessity and sufficiency tests. We start from the community-only intervention: in its six layers we ablate all MLP neurons except the community members and the ``always on'' neurons. We then evaluate all $2^6=64$ subsets of retained community members, keeping the ``always on'' neurons always active. For each subset, we measure the mean KL divergence from the clean next-token distribution.

\texttt{L14:11888} alone recovers most of the community's behavior: retaining it alone gives a KL divergence of 0.68 compared to 4.56 when all six community members are removed and 0.06 when all six are present. Adding \texttt{L5:3846} reduces KL further to 0.07, producing almost identical behavior to the full community. \texttt{L5:3846} is not a unique partner for \texttt{L14:11888}: among the two-neuron subsets (four members removed), four of the five subsets containing \texttt{L14:11888} achieve KL below 0.5 (\Cref{fig:lattice}a). Several neurons can therefore complement its contribution.

This community shows strong redundancy across its member neurons. Despite its strong effect in isolation, removing \texttt{L14:11888} from the full community barely changes the prediction. Indeed, all six single-neuron ablations leave mean KL between 0.06 and 0.07, showing that the community tolerates losing any one member on these prompts (\Cref{fig:lattice}a). Larger removals reveal differences between members: removing both \texttt{L14:11888} and \texttt{L5:3846} gives a KL divergence of 0.65, and additionally removing \texttt{L12:14182} raises it to 2.20 (\Cref{fig:lattice}b).

Compensating for the loss of \texttt{L14:11888} does not require all five other members. Retaining just \texttt{L5:3846} and \texttt{L12:14182} gives a KL divergence of approximately 0.2, corresponding to a point without a black border at four members removed. This pair matches or improves on pairs combining \texttt{L14:11888} with \texttt{L2:5664}, \texttt{L12:14182}, or \texttt{L13:3604}. Thus, the strongest neuron in isolation is replaceable even within a small subset of the community.

For this community, the results suggest a form of \emph{redundant voting}: neurons make unequal but partly substitutable contributions to a common behavior. Multiple combinations preserve the prediction, allowing the community to tolerate individual failures even though removing several complementary members can substantially disrupt it. This also shows why necessity of a community as a group does not imply necessity of each member.

The degree of redundancy varies across communities, as the additional subset ablations in \Cref{app:community-subsets} show. In the in-context copying community, removing \texttt{L23:6134} alone raises mean KL divergence from 0.05 to 7.37, showing that a single member can be critical to a community's effect.

%% file: sections/8_discussion.tex
\textbf{Neuron-sparse behaviors.} Although we do not investigate why training produces neuron-sparse representations for some behaviors but not others, our results suggest a possible explanation. Allocating a small neuron group to a particular function carries a capacity cost, which could be offset by benefits for frequently used computations requiring precise answers, such as grammatical agreement, list formatting, or closing parentheses. We speculate that concentrating these computations in small neuron groups may help reduce interference from other functions. Our experiments do not directly test this explanation, and frequent behaviors are also easier for our pipeline to discover.

\begin{figure}[t]
    \centering
    \includegraphics[width=0.95\textwidth]{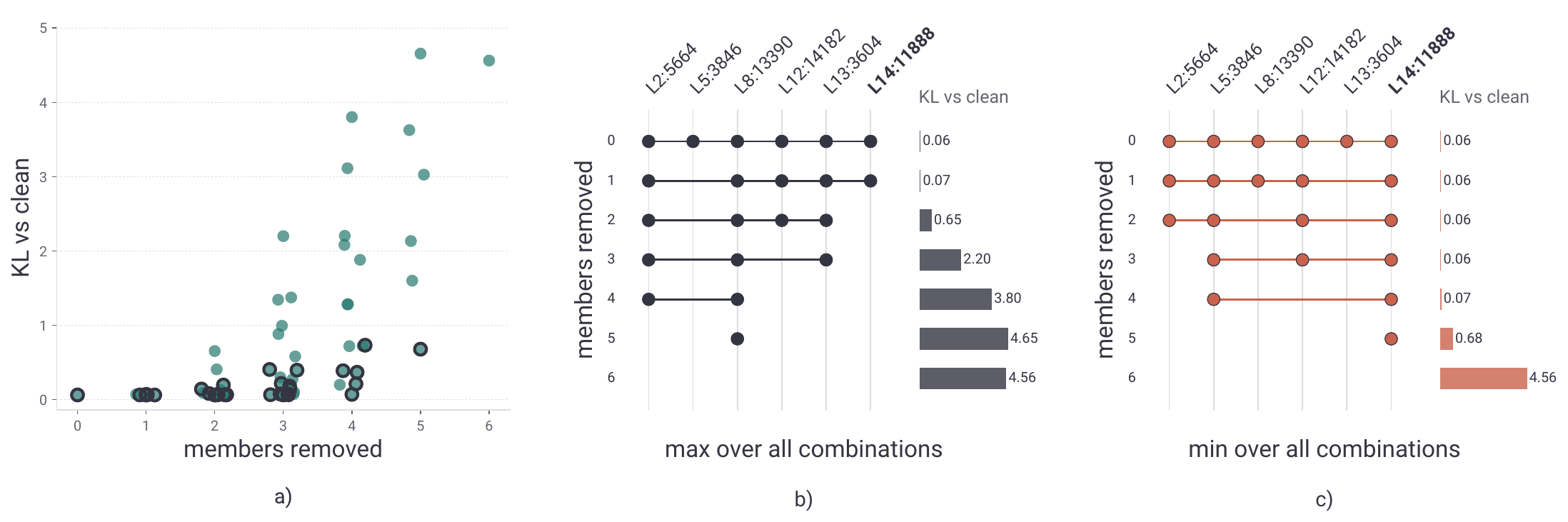}
    \caption{\textbf{Partial redundancy within the \emph{an} + vowel community.} Starting from the community-only intervention, we ablate every subset of its six members. \textbf{(a)} shows mean KL divergence for all possible combinations of removed member neurons. Black outlines mark runs where \texttt{L14:11888} is retained. \textbf{(b, c)} show subsets with the highest and lowest KL at each number of members removed. Dots indicate retained neurons.}
    \label{fig:lattice}
\end{figure}

\textbf{Behavior recall and interpretation.} Measuring recall requires defining a behavior and testing diverse prompts to distinguish broad coverage from narrow template matching. Unlike \emph{an} + vowel agreement, many behaviors have unclear boundaries and require manual inspection that is hard to scale. Automated interpretation may overlook inconsistencies; \citet{huang2023rigorously} document observational and causal errors in generated neuron explanations. Our coverage measurements (\Cref{sec:prompt-coverage}) therefore provide quantitative evidence for neuron-sparse behaviors, while the taxonomy (\Cref{sec:coverage-taxonomy}) only illustrates possible functions: its labels may be inaccurate or overly broad.

\textbf{Undiscovered behaviors.} Failure to discover a community does not establish that a behavior lacks a neuron-sparse mechanism. A mechanism may be missed because the discovery data contain too few relevant prompts, because attribution does not reliably identify its neurons, or because graph partitioning fails to group them together. We find community discovery to be sensitive to the data used to construct the graph, as well as to the resolution threshold. Our measured coverage therefore describes the communities recovered by this pipeline on the evaluation distribution. It does not establish an upper limit on how much of the model's behavior admits neuron-sparse explanations.

\textbf{Final token position.} We restrict attribution and interventions to the final prompt position only. This will inevitably miss communities that operate across multiple positions. For instance, in the arithmetic example "2+5=", some neurons might read the operands at earlier positions, and other neurons would read it at the final position to perform the addition. The current version of our method would not be able to find these connections. We see this approach as a reasonable first step, which allowed us to ignore the complexities associated with cross-position communities; we believe extending community discovery to cross-position setup is an important direction for future research.

\textbf{Non-overlapping communities.} At each resolution threshold, Leiden partitions the graph into disjoint communities, assigning each neuron to only one group. A neuron may, however, participate in several mechanisms, so this constraint need not reflect the model's organization. It can separate neurons that work together or force a shared neuron into only one of the groups that use it. Pooling communities across resolutions allows some overlap between candidate groups, but each individual partition retains this restriction. Allowing overlapping communities is a natural direction for extending the discovery method.

\textbf{Internal organization of communities.} Understanding how neuron community members work together is an important direction for future work. The subset ablations in \Cref{sec:deepdive} provide an initial view: the \emph{an} + vowel community contains several small combinations that preserve its effect, consistent with redundant voting. We expect discovered communities to represent different forms of functional objects. Some may form circuits in which distinct neurons perform successive computational steps, while others may repeatedly reinforce one or two shared directions across layers, providing redundancy around a simpler computation. Community discovery provides candidate groups for these analyses. The ablations in \Cref{sec:deepdive} are only the first step toward characterizing their mechanisms.

%% file: sections/appendix_arithmetic.tex
\citet{feucht2026arithmetic} identified 28 MLP neurons in layer 18 of Llama-3.1-8B that implement a base-10 addition mechanism reused across numerical addition and questions about hours, months, and weekdays. Starting from predefined arithmetic tasks, they use distributed alignment search to learn causal subspaces, then select neurons whose output weights align strongly with the learned output subspaces. We hypothesize that many behaviors in language models are similarly \emph{neuron-sparse}, i.e. they are mediated by small groups of neurons. In this work we aim to develop a method that can discover such groups at scale, without specifying the behaviors in advance. As a first test, we ask whether inexpensive gradient-based attribution can recover the neurons selected by this subspace analysis.

We evaluate 256 prompts for each of four templates: numerical addition, hours, months, and weekdays, using prompts from the released \emph{Arithmetic in the Wild} dataset. At the final prompt position, we score each neuron by $\left|h_j\,\frac{\partial\log p(\hat y\mid x)}{\partial h_j}\right|$, where $h_j$ is its activation and $\hat y$ is the model's highest-probability next token. This score requires a forward pass and a backward pass and estimates the magnitude of the neuron's effect on the prediction.

\begin{figure}[t]
    \centering
    \includegraphics[width=0.95\textwidth]{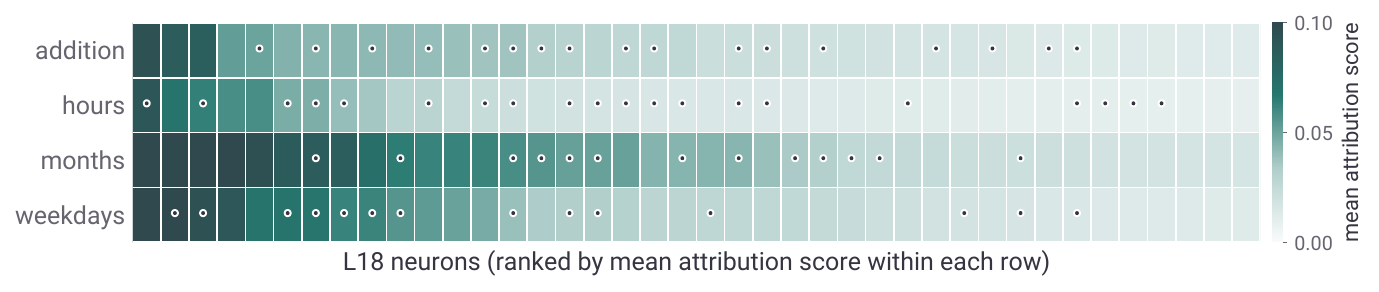}
    \caption{\textbf{Gradient-based attribution recovers known arithmetic neurons.} Each row shows the top 40 layer-18 neurons with the highest mean attribution score, ranked separately within the template. Black dots mark addition circuit neurons identified by \citet{feucht2026arithmetic}.}
    \label{fig:arithmetic-attribution}
\end{figure}

\Cref{fig:arithmetic-attribution} shows the top 40 neurons ranked by mean attribution for each template, with known neurons marked by black dots. The gradient-based attribution recovered 17/28 (60.7\%) reference neurons for addition, 20/26 (76.9\%) for hours, 13/16 (81.3\%) for months, and 14/15 (93.3\%) for weekdays. This recovery motivates using gradient attribution to identify candidate neuron groups across a broader corpus, followed by direct causal tests.

%% file: sections/appendix_attribution_ablation.tex
\begin{figure}
    \centering
    \includegraphics[width=\textwidth]{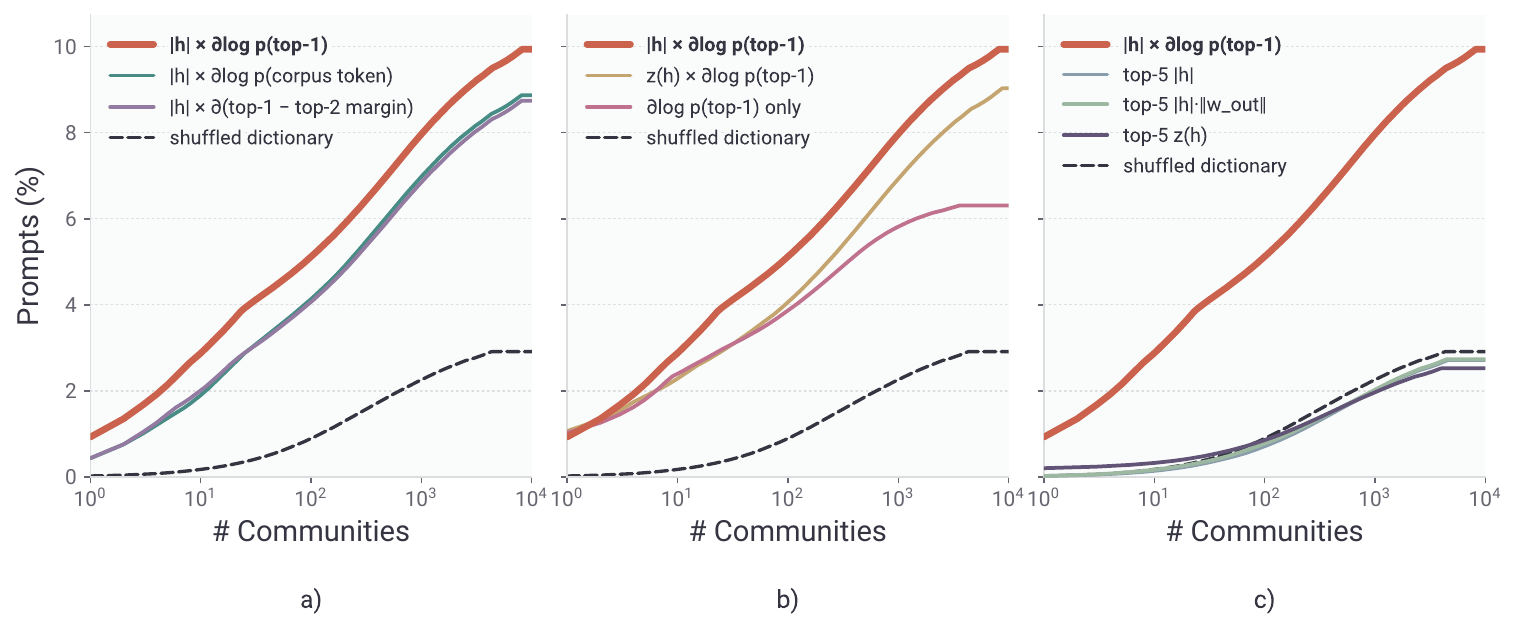}
    \caption{\textbf{Comparing neuron attribution methods.} Curves show the percentage of unique prompts explained as communities are added in greedy coverage order. The thick orange curve repeats the main paper's activation--gradient score in all panels; the dashed gray curve shows a baseline formed by shuffling neuron assignments between communities within each layer. \textbf{(a)} Alternative gradient targets. \textbf{(b)} Alternative activation factors. \textbf{(c)} Scores using activations without gradients. The horizontal axis is logarithmic and displays the first 10,000 communities. All gradient-based scores use gradient magnitudes.}
    \label{fig:attribution-ablation}
\end{figure}

We compare eight neuron-scoring methods and a shuffled-community baseline in a separate experiment at the one-million-prompt scale (\Cref{fig:attribution-ablation}). Each scoring method selects the top five neurons per layer at the prediction position. We discover communities over L0--L29, retain groups of 2--128 neurons, and pool and deduplicate them across the tested Leiden resolution thresholds. We measure coverage on the same 852,263 FineWeb-Edu, Wikipedia, and FineMath prompts shared by all runs. Communities are added greedily: each next community explains the largest number of previously unexplained prompts. All eight scoring methods and the shuffled-community baseline use the same three intervention tests and thresholds: a prompt is counted once if a community passes community necessity and layer necessity ($D_{\mathrm{KL}}\geq0.7$) and sufficiency ($D_{\mathrm{KL}}\leq0.2$), with all three interventions applied within the same tested layer subset.

All runs in this comparison also use the same set of ``always on'' neurons. Within the tested layer subset, sufficiency retains the community together with these always-on neurons, while layer necessity retains only the always-on neurons. Community necessity ablates only the community members in that subset. These rules are identical across all scoring methods and the shuffled baseline.

Panel (a) keeps the activation factor $|h|$ and changes the differentiated quantity: the log probability of the model's top prediction, the log probability of the next corpus token, or the log-probability margin between the top two predictions. Panel (b) keeps the gradient of the top prediction's log probability and replaces $|h|$ with either the absolute standardized activation, denoted $z(h)$, or a constant factor, giving a gradient-only score. Panel (c) removes gradients and ranks neurons by $|h|$, by $|h|\,\|w_{\mathrm{out}}\|_2$ (activation magnitude weighted by the neuron's output-vector norm), or by $z(h)$.

For the random baseline, we take the reference communities that explain at least one prompt and randomly permute their neuron assignments across communities within each layer. This preserves each community's layer set, size, and each neuron's frequency. We evaluate the shuffled communities using the same prompts, top-five attribution sets, candidate-prompt selection, causal tests, and greedy coverage ordering as the reference.

Within the first 10,000 communities shown in \Cref{fig:attribution-ablation}, our activation--gradient score achieves the highest coverage, at 9.94\%. Using the corpus-token or margin gradient yields 8.88\% and 8.74\%, respectively. Standardizing activations before multiplying by the gradient yields 9.04\%, while gradients alone yield 6.31\%. Activation-only scores achieve 2.72\% for activation magnitude, 2.74\% with output-norm weighting, and 2.53\% for standardized activations. These results support using both activation magnitude and sensitivity to the model's prediction to select neurons for discovery.

The shuffled dictionary covers less than 0.2\% of prompts with ten communities and reaches 2.91\% within the plotted range, compared with 9.94\% for the reference dictionary. Its final coverage slightly exceeds that of all three activation-only scores, but remains substantially below every gradient-based score.

%% file: sections/appendix_gamma_sweep.tex
We examine how many Leiden resolution thresholds are needed to retain most of the prompt coverage obtained from a broader sweep. We start with communities discovered at 14 geometrically spaced thresholds, $\gamma\in\{0.25,0.5,1,2,4,8,16,32,64,128,256,512,1024,2048\}$, and use the same 8,862,662 held-out prompts retained after deduplication and the same causal evaluation as in \Cref{sec:prompt-coverage}. For each threshold, we collect the prompts explained by at least one of its communities on a tested layer subset. Coverage for a subset of thresholds is the union of these prompt sets, counting each prompt once.

\begin{figure}
    \centering
    \includegraphics[width=0.7\textwidth]{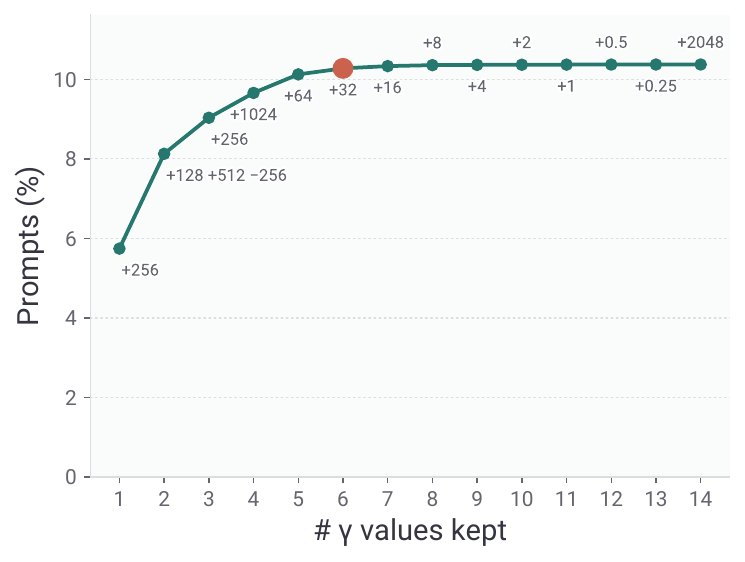}
    \caption{\textbf{Six resolution thresholds retain 99.1\% of the full sweep's coverage.} Each point shows the maximum prompt coverage achievable with the indicated number of thresholds, found by exhaustive subset search. Labels indicate thresholds added ($+$) or removed ($-$) relative to the optimal subset at the preceding point. The orange point marks the six thresholds used in the paper, $\gamma\in\{32,64,128,256,512,1024\}$, which cover 10.28\% of prompts, compared with 10.37\% for all 14 thresholds.}
    \label{fig:gamma-sweep}
\end{figure}

We enumerate all $2^{14}=16{,}384$ subsets and find the subset with the greatest coverage for each number of retained thresholds (\Cref{fig:gamma-sweep}). The best single threshold is $\gamma=256$, covering 5.74\% of prompts. The best pair, $\{128,512\}$, covers 8.13\%, illustrating that optimal subsets need not contain the best smaller subset. Adding $\gamma=256$ gives the optimal three-threshold subset, with 9.03\% coverage. Adding $\gamma=1024$ and then $\gamma=64$ increases coverage to 9.66\% and 10.13\%, respectively.

The six thresholds used in the main experiments, $\{32,64,128,256,512,1024\}$, form the optimal six-threshold subset on these prompts and cover 10.28\%. All 14 thresholds together cover 10.37\%, so the selected subset retains 99.1\% of the full union while using fewer than half the thresholds. Adding $\gamma=16$ raises coverage to 10.33\%; the remaining seven thresholds together add only about 0.04 percentage points. This plateau motivates the six-threshold choice as a compact set that preserves nearly all of the observed coverage. For the sweep above, subset selection and coverage measurement use the same evaluation prompts.

As an independent check, we repeat the search for the best six-threshold subset using only training prompts. This selects the same set, $\{32,64,128,256,512,1024\}$. \Cref{tab:gamma-train-test} compares coverage on the train data used to build communities, and on the held-out prompts used above. The coverage percentage is very similar across all sets of thresholds, showing that threshold selection process is not sensitive to a specific set of prompts, and translates well between train and held-out data.

\begin{table}[htbp]
    \centering
    \small
    \setlength{\tabcolsep}{5pt}
    \renewcommand{\arraystretch}{1.1}
    \caption{\textbf{Resolution coverage on training and held-out prompts.} Rows show the resolution sets from the original sweep. From the fourth row onward, $\{\ldots\}+\gamma$ adds the indicated resolution to the previous row's set. The six resolutions used in the main experiments, also recovered by training-only selection, are shown in bold.}
    \label{tab:gamma-train-test}
    \begin{tabular}{@{}lrr@{}}
        \toprule
        $\gamma$ values & \shortstack{Train\\coverage (\%)} & \shortstack{Held-out\\coverage (\%)} \\
        \midrule
        $\{256\}$ & 5.70 & 5.74 \\
        $\{128, 512\}$ & 8.05 & 8.13 \\
        $\{128, 256, 512\}$ & 8.98 & 9.03 \\
        $\{\ldots\} + 1024$ & 9.64 & 9.66 \\
        $\{\ldots\} + 64$ & 10.09 & 10.13 \\
        \textbf{\{\ldots\} + 32} & \textbf{10.21} & \textbf{10.28} \\
        $\{\ldots\} + 16$ & 10.26 & 10.33 \\
        $\{\ldots\} + 8$ & 10.29 & 10.36 \\
        $\{\ldots\} + 4$ & 10.30 & 10.36 \\
        $\{\ldots\} + 2$ & 10.30 & 10.37 \\
        $\{\ldots\} + 1$ & 10.30 & 10.37 \\
        $\{\ldots\} + 0.5$ & 10.30 & 10.37 \\
        $\{\ldots\} + 0.25$ & 10.30 & 10.37 \\
        $\{\ldots\} + 2048$ & 10.31 & 10.37 \\
        \bottomrule
    \end{tabular}
\end{table}

%% file: sections/appendix_data.tex
We augment the base corpora with five targeted data slices. \Cref{tab:targeted-data} describes their target positions and reports sampled and retained counts for the 10M training configuration and held-out evaluation. These counts cover only the targeted additions; the memorized-text slice has no held-out examples.

\begin{table}[htbp]
    \centering
    \footnotesize
    \setlength{\tabcolsep}{4pt}
    \renewcommand{\arraystretch}{1.15}
    \begin{tabularx}{\textwidth}{@{}lLrrrr@{}}
        \toprule
        & & \multicolumn{2}{c}{10M train} & \multicolumn{2}{c}{Held-out} \\
        \cmidrule(lr){3-4}\cmidrule(l){5-6}
        Slice & Target position & Sampled & Kept & Sampled & Kept \\
        \midrule
        Grammar & Word-initial closed-class token (determiner, preposition, or pronoun) in French, German, Spanish, or Russian text.
            & 100,000 & 99,000 & 100,000 & 99,000 \\
        \addlinespace
        Switch & First token after a language switch: Europarl~\citep{koehn2005europarl} bilingual pairs in four layouts, plus Wikipedia ``(Language: \ldots)'' glosses.
            & 29,000 & 29,000 & 29,000 & 25,000 \\
        \addlinespace
        Code & Prompts sampled at uniformly chosen token locations in source files from \texttt{the-stack-smol}~\citep{kocetkov2022stack}, covering 16 languages.
            & 261,000 & 246,000 & 261,000 & 243,000 \\
        \addlinespace
        QA & First answer token of the last question in a 5-shot quiz document (TriviaQA~\citep{joshi2017triviaqa} and NQ-open~\citep{kwiatkowski2019naturalquestions,lee2019latent}, in five formats).
            & 50,000 & 50,000 & 50,000 & 49,000 \\
        \addlinespace
        Memorized & Hits on famous public-domain text with $p>0.9$ where a 12-token context fails, plus controls from the same windows.
            & 48,000 & 46,000 & --- & --- \\
        \midrule
        Total & & 489,000 & 470,000 & 440,000 & 416,000 \\
        \bottomrule
    \end{tabularx}
    \caption{Targeted data slices in the 10M training configuration and held-out evaluation. Counts are shown before (sampled) and after (kept) filtering and deduplication, rounded to the nearest thousand. Totals are rounded independently from the unrounded counts.}
    \label{tab:targeted-data}
\end{table}

%% file: sections/appendix_taxonomy_prompt.tex
We reproduce the classification and description prompts for the 5,281 communities at $\gamma=256$ that explain at least ten held-out prompts. Classification used \texttt{gpt-5.6-terra} at temperature zero, with one request per community and up to 100 examples, without enrichment oversampling.

\subsection{Full taxonomy}
\input{sections/appendix_taxonomy_table.tex}

\begingroup
\fontencoding{T1}\selectfont
\fvset{fontsize=\scriptsize,breaklines=true,breakanywhere=true,frame=single,framesep=2mm}

\subsection{Classification system prompt}
\begin{Verbatim}
You classify a "neuron community" of Llama-3.1-8B by the FUNCTION it performs at the next-token prediction it controls. You get the community's size and layers and up to 100 real examples: the text before the target (ending where the model predicts the next token), the actual next token, the model's top predictions with probabilities, and the top predictions after the community's neurons are removed at that position ("removed"). The community is causally necessary and sufficient on these examples. Look at what changes between the intact and removed predictions, not at the topic of the text.

Assign exactly one PRIMARY label (the number), an optional SECONDARY label when a second reading is genuinely plausible, and a short lowercase FUNCTION slug naming the mechanism so that identical mechanisms get identical slugs across communities.

LABELS
 1 fixed phrase / collocation — one lexical trigger completes ONE fixed multiword expression or idiom tail ("each other", "from time to time", "take into account", "on behalf of"). *The completed string is the same across examples.*
 2 specific token promoted after a trigger family — one output token promoted after a family of related triggers (" of" after ability/protection, " in" after increase/rise, "along" after come/happen, "exhibit" in museum contexts). *Same output token, different triggers, and the trigger motivates it.*
 3 suppression of a default word — the community holds down one ordinary, topically plausible word or word family; when it is removed the prediction collapses onto that word in nearly every example, even where it is wrong (salt, budget, trust, bacteria, engine, wise, cup). *Look at the REMOVED column: it repeats one token at 0.6–1.0 across unrelated contexts.*
 4 local grammar (closed-set choice) — syntactic choice decided by adjacent words: a/an, there is/are, do-support, subject–auxiliary inversion, article after a preposition, which/that, imperative after a fronted clause. *Antecedent is within a few tokens.*
 5 long-range agreement / anaphora — form fixed by an earlier noun phrase or discourse referent: pronoun case/gender/number, possessive determiners (your/their/his), reflexives, copula number with an inverted or distant subject, VP-ellipsis auxiliary echo, polarity items licensed a clause earlier. *Antecedent is a sentence or more away.*
 6 trigger lemma selects its complement — one recurring trigger word or lemma, several acceptable continuations chosen by local context: verb/noun argument structure (demand → of/for/to; "within" → the/a/weeks), sense-dependent continuation of a polysemous word (bank, interest, stress), compound heads after a fixed modifier (full-, run-). *Same trigger, output token varies.*
 7 semantic content word / near-synonym choice — a specific content word required by sentence meaning, or the choice among a small set of near-synonyms for one role (" World" before War, "dedicated" for a church, victim/subject/target, end/goal/aim). *One lemma or one semantic role per community; the word is demanded, not merely on-topic.*
 8 number / arithmetic / units — computed or scale-dependent numeric output: ordinal suffix from digits, digit continuation to the right precision, base conversion, exponent/sign continuation, Roman numerals, unit of the right magnitude, era markers, multiplicative comparison
 9 punctuation / delimiter / layout in prose — delimiter and layout tokens in ordinary text: closing parenthesis/quote balance, what follows an opening bracket, resuming a clause after a parenthetical, list commas and dashes, em-dash continuation, periods after abbreviations, en dash vs hyphen in ranges, casing style
10 semantic formatting — structured values whose content matters: dates and citation digits, price/currency numerals, postcodes, phone numbers, time zones, e-mail local parts, call-sign suffixes, credentials after a name, the `@` sigil in handles
11 morphology / inflection choice — inflection or derivational variant of a word already started or determined: -ed vs -ing, plural -s/-es, possessive 's, -ic/-ical/-ics, -ise/-ize, -driven/-based, generational suffix (Jr., III). *Alternatives are genuine variants of one word.*
12 stem completion of common words — finishing a split common word or Latinate root into whichever word follows (arch- → archives/archaic, dis-, aut(o)-, fresh-, contribut-, colo(u)r-), contractions (didn → 't). *The fragment is a common-word stem and the completions are unrelated whole words.*
13 in-context copy — repeating a name, acronym, term or attribute that appears earlier in the prompt (vocatives, acronym expansion, title copy, a Wikipedia category attribute repeated from an adjacent line). *The string is present nearby.*
14 enumeration / sequence continuation — next item of an established series: next letter or number label, ordinal computed from listed items, ellipsis after a pattern, sequel numbering
15 code & markup syntax — programming/markup tokens governed by earlier code, including whitespace inside code: comment markers (// # /*), `$` and `&` sigils and operators, `()` on calls, closing HTML tag, type after `:`/`->`, SQL keywords, print statement of the language, LaTeX commands and brace balance, indentation width and column alignment in source, CLI subcommands, license headers inside code comments
16 non-English-specific grammar or script — mechanisms with no English counterpart: case-governed articles/pronouns/clitics (den/der, ihm/ihn, le/lo), governed prepositions in other languages (Spanish personal "a", Russian "при", French "à"), native script after a language label, the copula/connective/addressing formula opening a translated sentence (Es, Deshalb, Bien que, Señor Presidente). *If the same mechanism exists in English (e.g. lead→to / к / 到, pronoun agreement in Spanish), label the MECHANISM (4/5/6/…) and put 16 as secondary.*
17 factual / world-knowledge recall — fills an entity slot from parametric knowledge and the entity changes with the question: US presidents from a clue, composer of an opera, band or singer of a song, sports record holders, actor for a role, publisher/co-author in a citation, "Hurricane ___" names. *The answer is not in the context.*
18 topic / semantic-field vocabulary — promotes the vocabulary of one domain rather than a specific token or role: drinks, waste, monasteries, football leagues, ophthalmology, abstract algebra, a region's place names. *No single lemma dominates; removed predictions stay domain-plausible.* Last resort before 22/0.
19 context-independent token bias — pushes one token or short subword across contexts that share NO motivating trigger family, and is often wrong: function words (" the", " in", " of", "if", "by", "just"), first subwords (" D", " W", "wh", "Pre-"), digits ("0", "1", "18"), a default entity ("China"). *Two conditions, both required: the INTACT column shows the token even where wrong, AND the contexts have nothing in common that would call for it. A token over-applied inside its own family ("refer" in medical case handling, "high" in degree/scale phrases, "base" after customer/fan/home, "next" in sequencing phrases) is 2, not 19, even when it is sometimes wrong.*
20 proper-noun fragment completion — finishing a split name, place, acronym or taxon by orthography: letter/prefix hubs (B-, St-, L-, Sal-, Kn-), surname endings by script or region (Korean, Scandinavian, Basque, Hispanic "y", Hebrew, Indian, Sri Lankan), all-caps acronyms in parentheses, taxonomic binomials. *The fragment is a proper noun; nothing semantic is shared beyond the letters.*
21 template / boilerplate continuation — continuing a memorised document or site template rather than general prose structure: Wikipedia section headers and category lines ("Living people", "[Nationality] footballers", "-language films"), infobox and citation templates, forum and site chrome ("GMAT Club", "tutors are online", "Read Also"), copyright/license text in prose, the blank or leading-space token before a template entry. *The same near-verbatim template recurs across examples.*
22 trigger detector with heterogeneous effect — fires whenever one specific whole word or lemma appears (under/unter/под, format, name, volunteer, "3"), but the downstream tokens are neither completions of it nor its complements, and no single output pattern holds. *Use only after excluding 6, 12 and 20: if the outputs complete the trigger fragment into words or names (Red → giant/planet/Reda, Be → asts/atus/ings) it is 20 or 12; if they are the trigger word's own continuation slot (Future → spouses/injury/Studies, bank → of/established, hear → to/these) it is 6. "Varied outputs" alone is never enough for 22.* Distinct from 0 (no shared trigger either).
 0 other / unclear — no shared trigger and no shared effect after reading all examples (mixed content-word swaps, mixed digits/punctuation); describe what is seen

PRECEDENCE
1. **3 vs 19 vs 2 (which column carries the shared token?)** REMOVED repeats one token everywhere → 3. INTACT shows the token, and the
   contexts share a motivating trigger family (a domain, a construction, a collocation set) → 2, even if the token is sometimes wrong.
   INTACT shows the token with no motivating family at all → 19.
2. **1 vs 2 vs 6 (what is fixed?)** fixed multiword string → 1; fixed output token, varying triggers → 2; fixed trigger, varying
   output → 6.
3. **13 vs 17 vs 20/12 (where does the answer come from?)** present in the context → 13; from knowledge, changes with the question →
   17; the word is split mid-token and completed by orthography → 20 (proper noun) or 12 (common word).
Order when several fit: 3 (removed-column test), then the domain labels 15 (anything inside source or markup, incl. its whitespace), 8, 10,
21, 9, 20, 12, 11, 14, 13, 17; then grammar 5 over 4; then the lexical ladder 1 → 2 → 6 → 7; then 18; then 19 (only if no trigger family
motivates the token); then 22 (only if the outputs are neither completions nor complements of the trigger); 0 only when nothing fits.
16 is primary only for language-specific mechanisms (see its row); otherwise it is a secondary.
Expected base rates from the hand review of 1,378 communities: 19 and 22 are each about 5% and 1.5% of communities; if either exceeds
~10% of a batch, the tests above are being applied too loosely.

Answer with JSON only, no prose:
{"primary": <int>, "secondary": <int or null>, "function": "<slug>", "reason": "<one sentence>"}
\end{Verbatim}

\subsection{Classification community description}
Each user message contains the community size, layers, and number of prompts it explains, followed by up to 100 examples. Each example includes the last 200 characters of the prompt, the actual next token, and the top three predictions and their probabilities before and after community ablation.

\begin{Verbatim}
Community: 2 neurons, layers [0, 26] (subspan covering the most prompts: L26-L26); 21 held-out prompts explained by the community, 21 shown; prompt sources {'finemath': 20, 'code': 1}.
Each example: text tail ending where the next token is predicted ▸ | actual next token | intact top predictions (p) | removed top predictions (p)
  1. … homomorphism $$(\eta^*)^{q+k}\colon H^{q+k}(\Sigma^kM_+)\to H^{q+k}(X).$$⏎⏎Of course there is an isomorphism $$H^{q+k}(\Sigma^kM_+)=H^q(M)$$⏎⏎Next, we want to define a function $$\ker[(\eta^*)^{q+k}]▸ | actual " \\" | intact: " \\" 0.65, "$$" 0.18, "^" 0.06 | removed: "\\" 0.96, " \\" 0.02, "$$" 0.01
  2. …a}{\sqrt {16-\theta^2}}=\int \dfrac{d \theta}{\sqrt {4^2-\theta^2}}$ We know that $\int \dfrac{dx}{\sqrt {a^2-x^2}} dx=\sin^{-1} (\dfrac{x}{a})+C$; where $C$ is a constant of proportionality. Thus, $=▸ | actual " \\" | intact: " \\" 0.91, "sin" 0.07, " sin" 0.00 | removed: "\\" 0.60, " \\" 0.36, "sin" 0.03
  3. …her as $\varepsilon$ decreases, but otherwise, there is always at least one limit cycle. So my second question is:⏎⏎How would we show, analytically, that the system \begin{align*}x'&=y+x(\varepsilon+(▸ | actual "x" | intact: "x" 0.57, "1" 0.39, "2" 0.01 | removed: "\\" 0.68, "x" 0.19, "1" 0.12
  4. …\text{Exp}(\theta)$$ with $$\theta = \mathbb{E}[X]$$. The parameter space is $$\Omega = \{\theta \mid\theta>0\}.$$⏎⏎I need to contrast the next hypothesis: $$H_{0}:\theta \geq \theta_{0}$$ vs $$H_{1}:▸ | actual " \\" | intact: " \\" 0.96, "$$" 0.02, "0" 0.01 | removed: "\\" 0.81, " \\" 0.18, "$$" 0.00
  5. …is no constant term we can factor out an $$x$$ for $$x(x^2-3x-10)>0$$. We then plug into the quadratic formula for: $$x_1,x_2=\frac{-b \pm \sqrt{b^2 -4ac}}{2a}=\frac{3 \pm \sqrt{3^2 -4*1*(-10)}}{2*1}=▸ | actual "5" | intact: " \\" 0.29, "3" 0.20, "2" 0.12 | removed: "\\" 0.65, " \\" 0.10, "3" 0.07
  … (16 more examples, up to 100 per community)

Respond with the JSON classification only.
\end{Verbatim}

\subsection{Description system prompt}
The description request uses the same community evidence, ending with \texttt{Respond with the JSON description only.} in place of the classification instruction.

\begin{Verbatim}
You are analyzing a small group of neurons (a "community") inside Llama-3.1-8B. The examples are excerpts of WEB-SCRAPED text (Wikipedia-style articles, statistics pages, reports). Each example shows: text tail (the model predicts the next token at its end), actual continuation, top predictions, and top predictions after removing the community. The community is causally necessary and sufficient at these positions.
Write a short description. Respond with JSON only:
{"summary": "<one sentence: what the neuron group does>",
 "prompt": "<describe a typical prompt from the presented examples with enough detail to reproduce a similar text. Focus on what is shared by the presented examples. Note specific format, common tokens, and abbreviations. Make sure to include enough details to be able to produce more prompts like these>",
 "removal_effect": "<what happens to the prediction when the community is removed>"}
\end{Verbatim}
\endgroup

%% file: sections/appendix_taxonomy_table.tex
\begingroup
\footnotesize
\setlength{\tabcolsep}{4pt}
\renewcommand{\arraystretch}{1.12}
\begin{longtable}{@{}>{\raggedright\arraybackslash}p{0.31\textwidth}>{\raggedright\arraybackslash}p{\dimexpr0.69\textwidth-8pt\relax}@{}}
    \caption{Full behavioral taxonomy, grouped into the eight classes reported in \Cref{tab:coverage-taxonomy}. Definitions are condensed; the prompt below gives the complete decision and precedence rules.}
    \label{tab:full-taxonomy}\\
    \toprule
    \textbf{Label} & \textbf{Definition} \\
    \midrule
    \endfirsthead
    \multicolumn{2}{l}{\textbf{Table \thetable{} (continued)}}\\
    \toprule
    \textbf{Label} & \textbf{Definition} \\
    \midrule
    \endhead
    \midrule
    \multicolumn{2}{r}{\emph{Continued on the next page}}\\
    \endfoot
    \bottomrule
    \endlastfoot

    \multicolumn{2}{@{}l}{\textbf{Lexical / collocational}}\\*
    Fixed phrase / collocation & Completes the same fixed expression across examples, such as ``each other'' or ``take into account''. \\
    Specific token promoted after a trigger family & Promotes the same output token after different, related triggers that motivate it. \\
    Trigger lemma selects its complement & Selects different continuations after the same trigger word, according to context; for example, ``demand'' followed by ``of'', ``for'', or ``to''. \\
    Semantic content word / near-synonym choice & Selects a content word or near-synonym required by the sentence meaning, rather than merely related to its topic. \\
    \addlinespace

    \multicolumn{2}{@{}l}{\textbf{Default-token calibration}}\\*
    Suppression of a default word & Suppresses a word or word family that dominates predictions across unrelated contexts when the community is removed. \\
    Context-independent token bias & Promotes a token across contexts with no shared motivating trigger, including cases where that token is wrong. \\
    \addlinespace

    \multicolumn{2}{@{}l}{\textbf{Document, code, and format structure}}\\*
    Punctuation / delimiter / layout in prose & Controls punctuation, delimiter balance, capitalization, or layout in ordinary prose. \\
    Semantic formatting & Produces structured values whose content matters, such as dates, prices, postcodes, phone numbers, or citation digits. \\
    Enumeration / sequence continuation & Continues an established sequence, such as a list's next letter, number, or ordinal. \\
    Code \& markup syntax & Produces tokens governed by code or markup syntax, including operators, tags, commands, comments, and indentation. \\
    Template / boilerplate continuation & Continues a recurring document or site template, such as Wikipedia category lines, headers, or copyright text. \\
    \addlinespace

    \multicolumn{2}{@{}l}{\textbf{Grammar and agreement}}\\*
    Local grammar (closed-set choice) & Chooses a grammatical form using nearby words, such as article choice, auxiliary choice, or local agreement. \\
    Long-range agreement / anaphora & Chooses a form using a distant noun phrase or discourse referent, such as pronoun agreement or an auxiliary echo. \\
    Morphology / inflection choice & Chooses an inflectional or derivational variant of a word, such as plural \emph{-s} or \emph{-ed} versus \emph{-ing}. \\
    Non-English-specific grammar or script & Controls grammar or script without an English counterpart. Used as a secondary label when the mechanism also exists in English. \\
    \addlinespace

    \multicolumn{2}{@{}l}{\textbf{Sub-token / orthographic completion}}\\*
    Stem completion of common words & Completes a common-word stem or contraction; alternative completions can form different words. \\
    Proper-noun fragment completion & Completes a name, place, acronym, or taxon from its spelling, without requiring a shared semantic property. \\
    \addlinespace

    \multicolumn{2}{@{}l}{\textbf{Topic / semantic field}}\\*
    Topic / semantic-field vocabulary & Promotes vocabulary from a shared domain, without one dominant word or a specific required semantic role. \\
    \addlinespace

    \multicolumn{2}{@{}l}{\textbf{Retrieval and computation}}\\*
    Number / arithmetic / units & Produces computed or scale-dependent outputs, including arithmetic, numeric precision, ordinal suffixes, and units. \\
    In-context copy & Repeats a name, term, acronym, or attribute already present in the prompt. \\
    Factual / world-knowledge recall & Retrieves an entity from model knowledge; the answer varies with the question and is absent from the prompt. \\
    \addlinespace

    \multicolumn{2}{@{}l}{\textbf{Other}}\\*
    Trigger detector with heterogeneous effect & Responds to a shared word or lemma with varied effects that are neither fragment completions nor the trigger's own complements. \\
    Other / unclear & Has neither a shared trigger nor a shared effect across the examples. \\
\end{longtable}
\endgroup

%% file: sections/appendix_community_subsets.tex
We apply the subset-ablation procedure from \Cref{sec:deepdive} to three further communities from \Cref{tab:community-examples}. For each, we sample 100 prompts on which the community passes necessity and sufficiency and evaluate every subset of its members under the community-only intervention, retaining the ``always on'' neurons. Each point in panel (a) reports mean KL divergence from the clean prediction for one retained subset. Panels (b) and (c) show the subsets with the highest and lowest mean KL at each removal count; dots mark retained neurons, and rows are selected independently. Black outlines in panel (a) indicate that all neurons whose labels are bold remain present.

\begin{figure}[htbp]
    \centering
    \includegraphics[width=\textwidth]{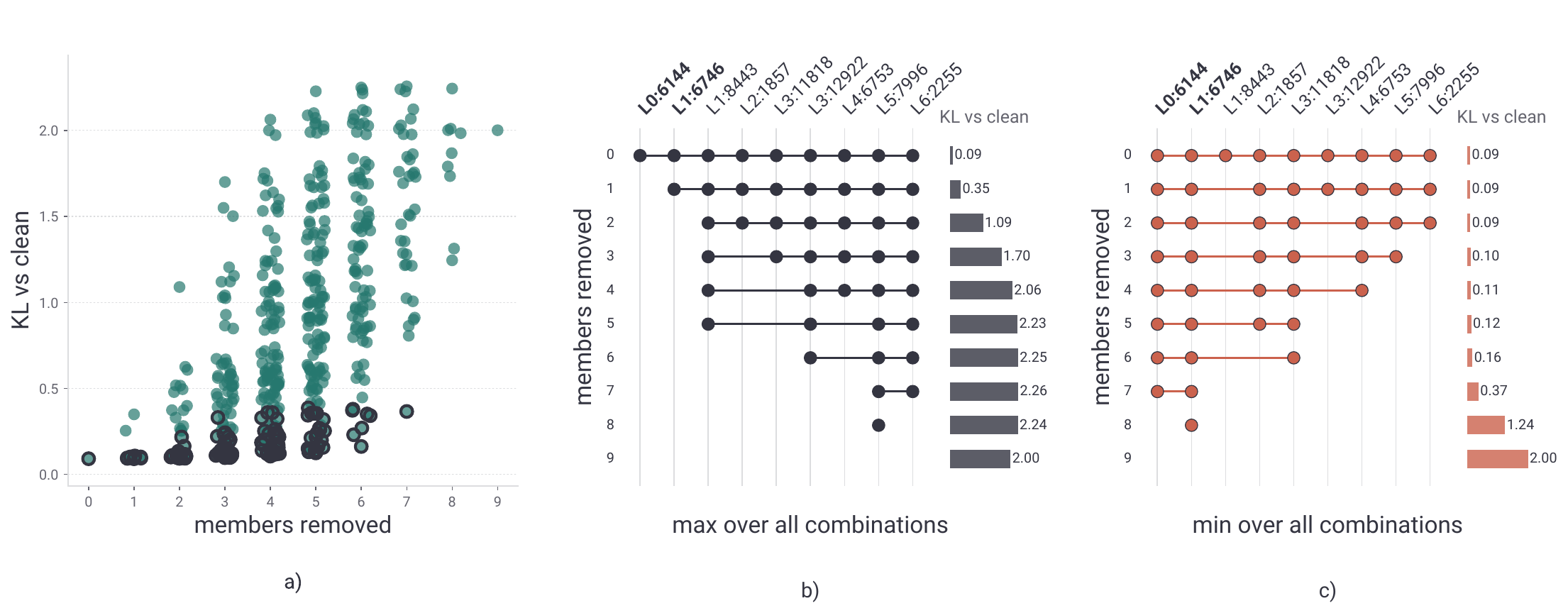}
    \caption{\textbf{Subset ablations for plural \emph{-s} morphology.} The tested community contains nine neurons across seven layers (L0--L6). Black outlines mark subsets retaining both \texttt{L0:6144} and \texttt{L1:6746}.}
    \label{fig:lattice-plural}
\end{figure}

\begin{figure}[htbp]
    \centering
    \includegraphics[width=\textwidth]{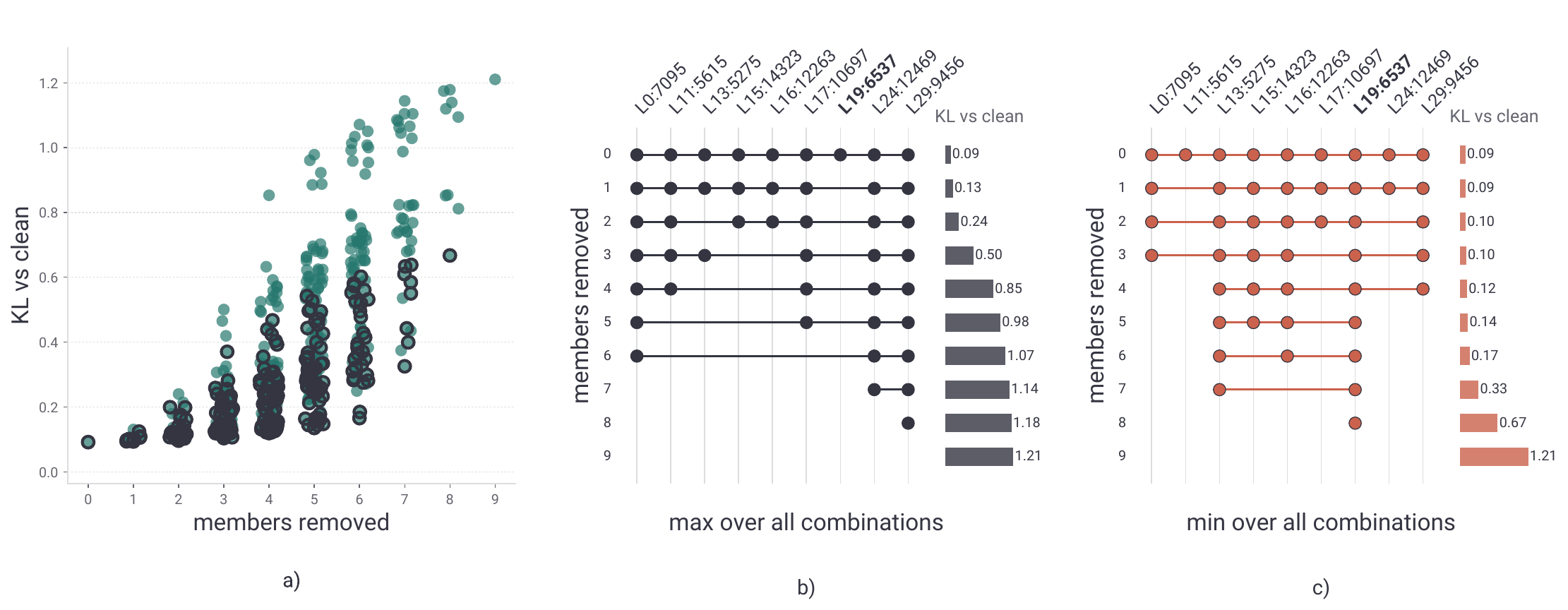}
    \caption{\textbf{Subset ablations for \emph{there is/are} agreement.} The tested community contains nine neurons across nine layers spanning L0--L29. Black outlines mark subsets retaining \texttt{L19:6537}.}
    \label{fig:lattice-there}
\end{figure}

\clearpage
\begin{figure}[ht]
    \centering
    \includegraphics[width=\textwidth]{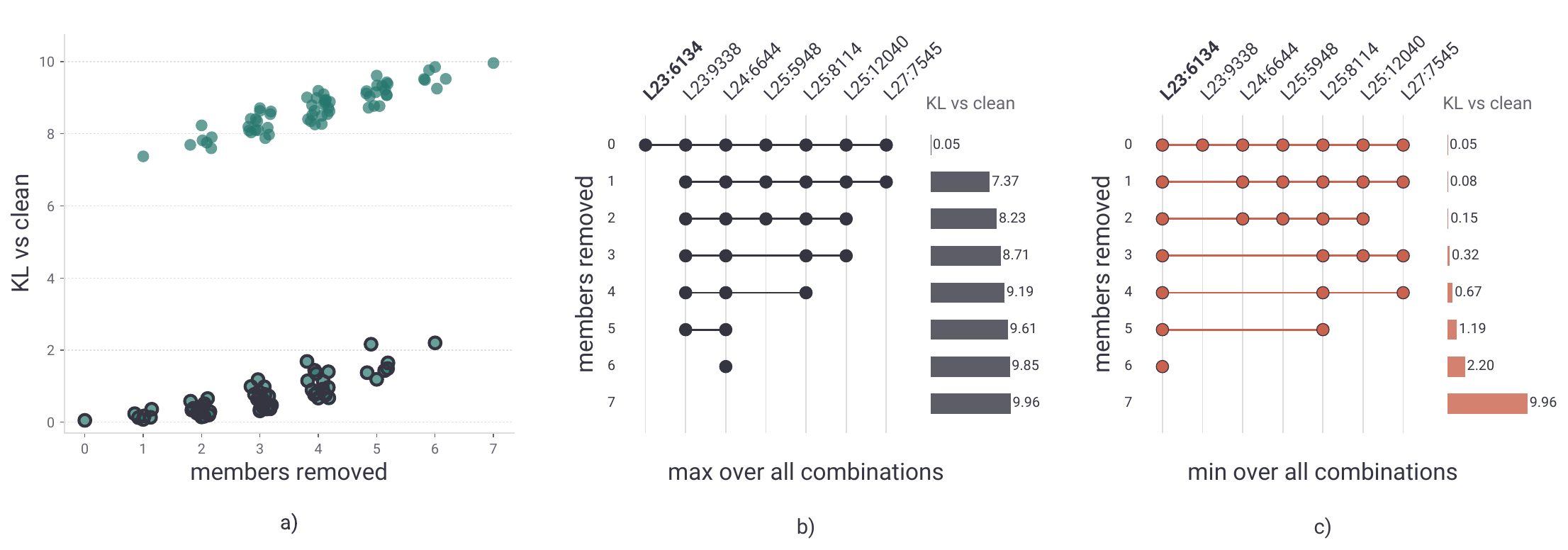}
    \caption{\textbf{Subset ablations for in-context copying.} The tested community contains seven neurons across four layers spanning L23--L27. Black outlines mark subsets retaining \texttt{L23:6134}.}
    \label{fig:lattice-copy}
\end{figure}